\documentclass[10pt,twocolumn,letterpaper]{article}

\usepackage{cvpr}              %

\usepackage{graphicx}
\usepackage{amsmath}
\usepackage{amssymb}
\usepackage{booktabs}
\usepackage{comment}

\usepackage[pagebackref,breaklinks,colorlinks]{hyperref}

\usepackage[capitalize]{cleveref}
\crefname{section}{Sec.}{Secs.}
\Crefname{section}{Section}{Sections}
\Crefname{table}{Table}{Tables}
\crefname{table}{Tab.}{Tabs.}

\newcommand{\ta}[1]{{\textcolor{red}{[\textbf{TA}: #1]}}}
\renewcommand{\ta}[1]{{\textcolor{red}{}}}
\renewcommand{\hat}{\widehat}
\renewcommand{\tilde}{\widetilde}

\usepackage{caption}
\usepackage{arydshln}
\usepackage{dsfont}
\usepackage{multirow}
\usepackage{rotating}
\usepackage[inline]{enumitem}
\newcommand*\diff{\mathop{}\!\mathrm{d}}

\def\cvprPaperID{****} %
\def\confName{CVPR}
\def\confYear{2023}

\begin{document}

\title{SPIn-NeRF: Multiview Segmentation and Perceptual Inpainting with Neural Radiance Fields}

\author{
Ashkan Mirzaei$^\text{1,2}$~~~~~~~~~~Tristan Aumentado-Armstrong$^\text{1,2,4}$~~~~~~~~~~~Konstantinos G. Derpanis$^\text{1,3,4}$ \\
Jonathan Kelly$^\text{2}$~~~~~~~~Marcus A. Brubaker$^\text{1,3,4}$~~~~~~~~Igor Gilitschenski$^\text{2}$~~~~~~~~Alex Levinshtein$^\text{1}$\\
$^\text{1}$Samsung AI Centre Toronto~~$^\text{2}$University of Toronto~~$^\text{3}$York University~~$^\text{4}$Vector Institute for AI\\
{\tt\small \{a.mirzaei,tristan.a\}@partner.samsung.com,~\{jkelly,gilitschenski\}@cs.toronto.edu}\\
{\tt\small \{kosta,mab\}@eecs.yorku.ca,~alex.lev@samsung.com}
}

\twocolumn[{%
\renewcommand\twocolumn[1][]{#1}%
\maketitle
\begin{center}
    \centering
    \captionsetup{type=figure}
    \includegraphics[width=.8\textwidth,width=0.99\linewidth]{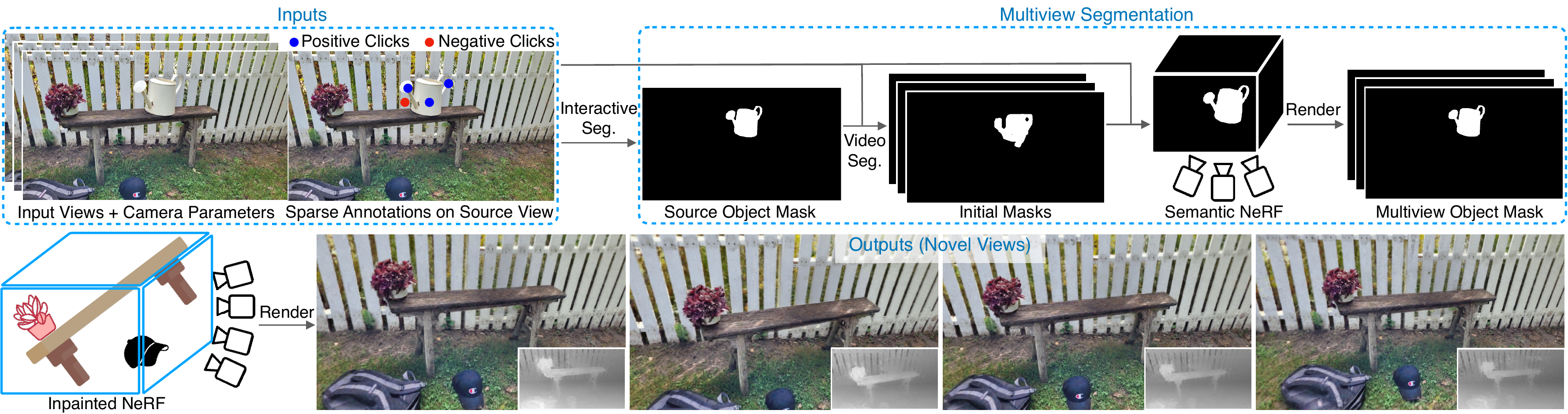}
    \captionof{figure}{An example of the inputs and outputs of our 3D inpainting framework. In addition to the images captured from the scene and their corresponding camera parameters, users are tasked with providing a few points in a single image to indicate which object they wish to remove from the scene (upper-left inset). These sparse annotations are then automatically transferred to all other views, and utilized for multiview mask construction (upper-right inset). 
    The resulting 3D-consistent mask is used in a perceptual optimization problem that results in 3D scene inpainting (lower row), with rendered depth from the optimized NeRF shown for each image as an inset.
    }
    \label{fig:banner}
\end{center}%
}]

\begin{abstract}
   Neural Radiance Fields (NeRFs) have emerged as a popular approach for novel view synthesis. While NeRFs are quickly being adapted for a wider set of applications, intuitively editing NeRF scenes 
is still an open challenge. One important editing task is the removal of unwanted objects from a 3D scene, such that the replaced region is visually plausible and consistent with its context. %
We refer to this task as \emph{3D inpainting}.
In 3D, solutions must be both consistent across multiple views and geometrically valid.
In this paper, we  propose a novel 3D inpainting method that addresses these challenges.
Given a small set of posed images and sparse annotations in a single input image, our framework first rapidly obtains a 3D segmentation mask for a target object. %
Using the mask, a perceptual optimization-based approach is then introduced that leverages learned 2D image inpainters, distilling their information into 3D space, while ensuring view consistency. 
We also address the lack of a diverse benchmark for evaluating 3D scene inpainting methods by introducing a dataset comprised of challenging real-world scenes.
In particular, our dataset contains views of the same scene with and without a target object, enabling more principled benchmarking of the 3D inpainting task.
We first demonstrate the superiority of our approach on multiview segmentation, comparing to NeRF-based methods and 2D segmentation approaches. We then evaluate on the task of 3D inpainting, establishing state-of-the-art performance against other NeRF manipulation algorithms, as well as a strong 2D image inpainter baseline.

\end{abstract}

\section{Introduction}
\label{sec:intro}

Neural rendering methods, especially Neural Radiance Fields (NeRFs)~\cite{original.nerf}, have recently emerged as a new modality for representing and reconstructing scenes~\cite{advances.in.neural.rendering}, achieving impressive results for novel view synthesis. Substantial research effort continues to focus on formulating more efficient NeRFs (e.g., \cite{mobile.nerf,rao2022icarus,Hu_2022_CVPR}), to make NeRFs more accessible in use-cases with more limited computational resources. As NeRFs become more widely accessible, the need for editing and manipulating the scenes represented by NeRFs will continue to grow. One notable editing application is removing objects and inpainting the 3D scene, analogous to the well-studied 2D image inpainting task~\cite{jam2021comprehensive}. 
However, several obstacles impede progress on this task, not only for the 3D inpainting process itself, but also in obtaining the input segmentation masks.
First, NeRF scenes are implicitly encoded within the neural mapping weights, resulting in an entangled and uninterpretable representation that is non-trivial to manipulate (compared to, say, the explicit discretized form of 2D image arrays or meshes in 3D).
Moreover, any attempt to inpaint a 3D scene must not only generate a perceptually realistic appearance in a single given view, but also preserve fundamental 3D properties, such as appearance consistency across views and geometric plausibility.
Finally, to obtain masks for the target object, it is more intuitive for most end users to interact with 2D images, rather than 3D interfaces; however, requiring annotations of multiple images (and maintaining view-consistent segments) is burdensome to users.
An appealing alternative is to expect only a minimal set of annotations for a single view.
This motivates a method capable of obtaining a view-consistent 3D segmentation mask of the object (for use in inpainting) from single-view sparse annotations.

In this paper, we address these challenges with an integrated method that takes in multiview images of a scene, extracts a 3D mask with minimal user input, and fits a NeRF to the masked images, such that the target object is replaced with plausible 3D appearance and geometry.
Existing interactive 2D segmentation methods do not consider the 3D aspects of the problem (e.g., \cite{ramadan2020survey}), while current NeRF-based approaches are unable to use sparse annotations~\cite{semantic.nerf} to perform well, or do not attain sufficient accuracy~\cite{neural.object.selection}.
Similarly, while some current NeRF manipulation algorithms allow object removal, 
they do not attempt to provide perceptually realistic inpaintings of newly unveiled parts of space (e.g., \cite{yang2021learning}).
To our knowledge, this is the first approach that handles both interactive multiview segmentation and full 3D inpainting in a single framework.

Our technique
leverages off-the-shelf, 3D-unaware models for segmentation and inpainting, and transfers their outputs to 3D space in a view-consistent manner. 
Building on the (2D) interactive segmentation~\cite{edge.flow,liu2021paddleseg,paddleseg2019} literature, our framework starts from a small number of user-defined image points on a target object (and a few negative samples outside it). %
From these, our algorithm initializes masks with a video-based model~\cite{dino}, and lifts them into a coherent 3D segmentation via fitting a semantic NeRF~\cite{semantic.nerf,ilabel,laterf}.
Then, after applying a pretrained 2D inpainter~\cite{lama} to the multiview image set, a customized NeRF fitting process is used to reconstruct the 3D inpainted scene, utilizing perceptual losses~\cite{perceptual} to account for inconsistencies in the 2D inpainted images, as well as inpainted depth images to regularize the geometry of the masked region.
Overall, we provide a complete method, from object selection to novel view synthesis of the inpainted scenes, in a unified framework with minimal burden on the user, illustrated in Figure~\ref{fig:banner}.

We demonstrate the effectiveness of our approach through extensive qualitative and quantitative evaluations. In addition, we address the lack of a benchmark for comparing scene inpainting methods, and introduce a new dataset where the ``ground-truth inpaintings'' (i.e., real images of the scene without the object) are available as well. 

In summary, our contributions are as follows:
    (i)
    a complete process for 3D scene manipulation, 
    starting from object selection with minimal user interaction and ending 
    with a 3D inpainted NeRF scene;
    (ii)
    to perform such selection,
    an extension of 2D segmentation models to the multiview case, %
    capable of recovering 3D-consistent masks from sparse annotations;
    (iii)
    to ensure view-consistency and perceptual plausibility, 
    a novel optimization-based formulation of 3D inpainting in NeRFs, which leverages 2D inpainters; 
    and
    (iv)
    a new dataset for 3D object removal evaluation that includes corresponding object-free ground-truth.

\section{Related Work}
\label{sec:related}

\textbf{Image Inpainting}. 
Inpainting in 2D %
has received significant research
attention~\cite{jam2021comprehensive}. 
While early techniques relied on patches~\cite{exemplar.based.inpaintint,million.photographs}, recent neural methods optimize both perceptual realism and reconstruction (e.g., \cite{lama,li2022mat,jain2022keys}). 
Various lines of research continue to be explored for improving visual fidelity,
including 
adversarial training (e.g., \cite{feature.learning.by.inpainting,comodgan}), 
architectural advances (e.g., \cite{shiftnet,li2022mat,coherent.sem.attention,lbam}), 
pluralistic outputs (e.g., \cite{comodgan,pluralistic}), 
multiscale processing (e.g., \cite{glob.local.completion,generative.inpainting,generative.multi.column.conv}),
and 
perceptual metrics (e.g., \cite{perceptual,lama,jain2022keys}).
Our work leverages LaMa~\cite{lama}, which applies frequency-domain transforms~\cite{ffc} inspired by transformers~\cite{vit,taming.vit}. %
Yet, none of these lift the problem into 3D;
thus, inpainting multiple captures of a scene in a consistent manner remains an underexplored task. 
While there are some existing 3D-aware image inpainting algorithms, 
they either only partially operate in 3D~\cite{yao20183d}, rely on reference images~\cite{zhao2022geofill}, or consider more limited scenarios~\cite{jampani2021slide}.
In contrast, our method operates directly in 3D, via the multiview-based NeRF model.

\textbf{NeRF Manipulation}.
Representing scenes via volumetric rendering has recently become an important research direction~\cite{advances.in.neural.rendering}. Based on differentiable volume rendering~\cite{henzler2019platonicgan,tulsiani2017mvsupervision} and positional encoding~\cite{gehring2017convolutional,vaswani2017attentionisallyouneed,tancik2020fourier}, NeRFs~\cite{original.nerf} have demonstrated remarkable performance in novel-view synthesis. Recent works have studied potential improvements in NeRF's training or rendering speed~\cite{plenoxels,tensorf,instant.ngp,plenoctrees,mobile.nerf,Hedman_2021_ICCV}, reconstruction quality~\cite{ds.nerf,mipnerf.360,mipnerf,bacon}, and data requirements~\cite{pixel.nerf,ibrnet,barf,wang2021nerf}. 
However, manipulating NeRF scenes remains a challenge. Clip-NeRF~\cite{clipnerf}, Object-NeRF~\cite{yang2021learning}, LaTeRF~\cite{laterf}, and others~\cite{nerf.editing,liu2021editing,conerf} introduce approaches to alter and complete objects represented by NeRFs; however, their performance is limited to simple objects, rather than scenes with significant clutter and texture, or they focus on tasks other than general inpainting (e.g., recoloring or deforming). 
Most closely related to our method is NeRF-In~\cite{nerf.in}, a concurrent unpublished work, which inpaints NeRF scenes with geometry and radiance priors from 2D image inpainters, but does not address inconsistency; instead, its use of a simple pixelwise loss relegates it to simply reducing the number of views used for fitting, which reduces final view synthesis quality.
Similarly using a pixelwise loss, the concurrent Remove-NeRF model~\cite{weder2022removing} reduces inconsistencies by excluding views based on an uncertainty mechanism.
In contrast, our approach is able to inpaint NeRF representations of challenging real-world scenes by incorporating 2D information in a view-consistent manner, via a \textit{perceptual}~\cite{perceptual} training regime.
This avoids over-constraints on the inpainting, which would normally lead to blurriness.

\section{Background: Neural Radiance Fields}
NeRFs~\cite{original.nerf} encode a 3D scene as a function, $f:(x, d) \rightarrow (c, \sigma)$, that maps a 3D coordinate, $x$, and a view direction, $d$, to a color, $c$, and density, $\sigma$. The function $f$ can be modelled in various ways (e.g., \cite{original.nerf,plenoxels}). 
For a ray, $r$, the expected color is estimated by volumetric rendering via quadrature; the ray is divided into $N$ sections between $t_n$ and $t_f$ (the near and far bounds), with $t_i$ sampled from the $i$-th section, to render the estimated color, $\hat{C}(r)$:
\begin{equation}
    \label{eq:volumetric.rendering.discrete}
    \hat{C}(r) = \sum_{i = 1}^{N} T_i(1 - \exp(-\sigma_i \delta_i))c_i,
\end{equation}
where $T_i = \exp(-\sum_{j = 1}^{i - 1} \sigma_j \delta_j)$ is the transmittance, $\delta_i = t_{i + 1} - t_i$ is the distance between two adjacent points, and $c_i$ and $\sigma_i$ are the color and density at $t_i$, respectively. For the rays passing through pixels of the training views, the ground-truth color $C_{\text{GT}}(r) $ is available, and the model is optimized using the reconstruction loss:
\begin{equation}
    \label{eq:reconstruction.loss}
    \mathcal{L}_\text{rec} = \sum_{r \in \mathcal{R}} \Vert \hat{C}(r) - C_{\text{GT}}(r) \Vert ^2,
\end{equation}
where $\mathcal{R}$ is a ray batch sampled from the training views. 

\section{Method}
\label{sec:method}

Given a set of RGB images, $\mathcal{I} = \{ I_i \}_{i= 1}^{n}$, with corresponding 3D poses, $\mathcal{G} = \{ G_i \}_{i= 1}^{n}$, and camera intrinsic matrix, $K$, our model expects one additional ``source'' view with sparse user annotations (i.e., a few points identifying the unwanted object).
From these inputs, we produce a NeRF model of the scene, capable of synthesizing an \textit{inpainted} image from any novel view.
We begin by obtaining an initial 3D mask from the single-view annotated source (\autoref{sec:mask.init}), followed by fitting a semantic NeRF, to improve the consistency and quality of the mask (\autoref{sec:mv.seg.nerf}).
Finally, in~\autoref{sec:mv.inpaint} we describe our view-consistent inpainting method, which takes the views and recovered masks as inputs. Our approach leverages the outputs of 2D inpainters~\cite{lama} as appearance and geometry priors to supervise the fitting of a new NeRF.
Figure~\ref{fig:banner} illustrates our entire approach, including the inputs and outputs. 
Additional details are in our supplementary material. 

\begin{figure}[t]
  \centering
   \includegraphics[width=0.99\linewidth]{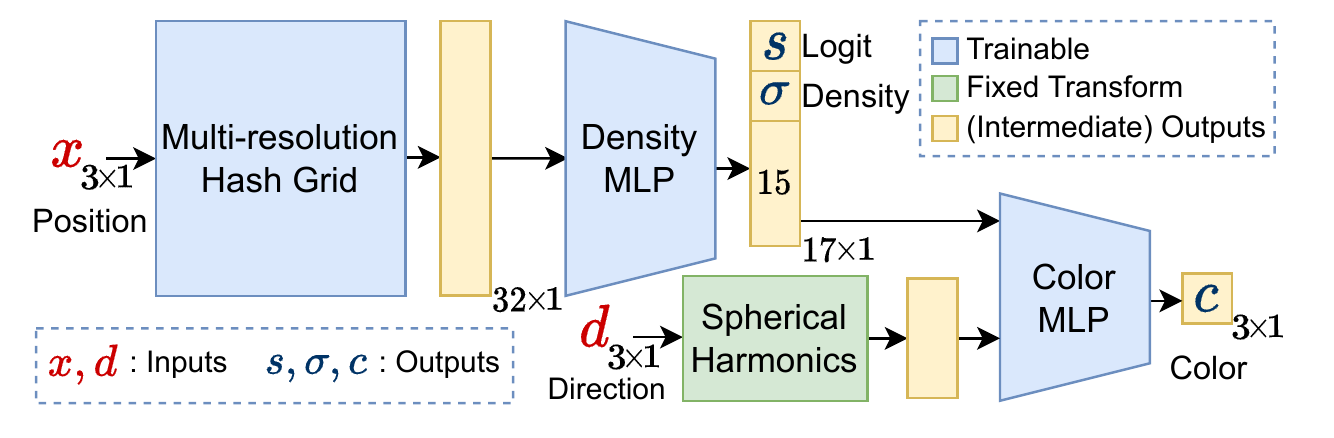}

   \caption{Overview of our multiview segmentation architecture. As input, this network takes in a 3D coordinate, $x$, and a view direction, $d$, and returns view-independent density, $\sigma(x)$, objectness logit, $s(x)$, and view-dependent color, $c(x, d)$. }
   \label{fig:mv.seg.architecture}
\end{figure}

\subsection{Multiview Segmentation}\label{sec:mv.seg}

\subsubsection{Mask Initialization}\label{sec:mask.init}
We first describe how we initialize a rough 3D mask from single-view annotations.
Denote the annotated source view as $I_1$.
The sparse information about the object and the source view are given to an interactive segmentation model~\cite{edge.flow} to estimate the initial source object mask, $\hat M_1$. The training views are then treated as a video sequence and, along with $\hat M_1$, given to a video instance segmentation model $V$~\cite{video.seg.survey,dino}, to compute
$ V(\{ I_i \}_{i= 1}^{n}, \hat M_1) = \{ \hat M_i \}_{i = 1}^{n}, $
where $\hat M_i$ is the initial guess for the object mask for $I_i$.
The initial masks, $\{ \hat M_i \}_{i = 1}^{n}$, are typically inaccurate around the boundaries, since the training views are not actually adjacent video frames, and video segmentation models are usually 3D-unaware. 
Hence, we use a semantic NeRF model~\cite{semantic.nerf,ilabel,laterf} to resolve the inconsistencies and improve the masks (\autoref{sec:mv.seg.nerf}), thus obtaining the masks for each input view, $\{M_i\}_{i=1}^{n}$, to use for inpainting (\autoref{sec:mv.inpaint}). 

\subsubsection{NeRF-based Segmentation}\label{sec:mv.seg.nerf}

Our multiview segmentation module takes the input RGB images, $\{I_i\}_{i=1}^{n}$, the corresponding camera intrinsic and extrinsic parameters, and the initial masks, $\{ \hat M_i \}_{i = 1}^{n}$, and trains a semantic NeRF~\cite{semantic.nerf}. Figure~\ref{fig:mv.seg.architecture} depicts the network used in the semantic NeRF; for a point, $x$, and a view direction, $d$, in addition to a density, $\sigma(x)$, and color, $c(x, d)$, it returns a pre-sigmoid ``objectness'' logit, $s(x)$. The objectness probability is then acquired as $p(x) = \text{Sigmoid}\big(s(x)\big)$. We use Instant-NGP~\cite{instant.ngp,tiny.cuda.nn,hash.nerf} as our NeRF architecture due to its fast convergence. The expected objectness logit, $\hat S(r)$, associated with a ray, $r$, is obtained by rendering the logits of the points on $r$ instead of their colors, with respect to the densities, in Eq.~\ref{eq:volumetric.rendering.discrete}~\cite{laterf}:
\begin{equation}
    \label{eq:logit.volumetric.rendering.discrete}
    \hat S(r) = \sum_{i=1}^{N} T_i (1 - \exp(-\sigma_i \delta_i)) s_i,
\end{equation}
where for simplicity, $s(r(t_i))$ is denoted by $s_i$. The objectness probability of a ray, $\hat P(r) = \text{Sigmoid}\big(\hat S(r)\big)$, is then supervised using the classification loss:
\begin{equation}
    \label{eq:clf.loss}
    \mathcal{L}_\text{clf} = \frac{1}{\vert \mathcal{R} \vert}\sum_{r \in \mathcal{R}} \text{BCE} \big(  \mathds{1}_{r \in \mathcal{R}_\text{masked}}, \hat P(r) \big),
\end{equation}
where $\mathds{1}$ is the indicator function, $\text{BCE}$ stands for the binary cross entropy loss, and $\mathcal{R}_\text{masked}$ is the set of rays passing through pixels that are masked in $\{ \hat M_i \}_{i = 1}^{n}$. 
During the calculation of the classification loss, $\mathcal{L}_\text{clf}$, the weights of the colors in the rendering equation (Eq.~\ref{eq:volumetric.rendering.discrete}) are detached to limit the supervised updates to the logits; this prevents changes to the existing geometry, due to gradient updates altering the $\sigma$ field. 
The geometry is supervised using a reconstruction loss, $\mathcal{L}_\text{rec}$, as in NeRF~\cite{original.nerf}, via the given RGB images. 
The overall loss, used to supervise the NeRF-based multiview segmentation model, is then given by:
\begin{equation}
    \label{eq:mvseg.loss}
    \mathcal{L}_\text{mv} = \mathcal{L}_\text{rec} + \lambda_\text{clf} \mathcal{L}_\text{clf},
\end{equation}
where the classification weight, $\lambda_\text{clf}$, is a hyper-parameter. 
After optimization, 3D-consistent masks, $\{ M_i \}_{i=1}^{n}$, are obtained by thresholding the objectness probabilities and masking the pixels with probabilities greater than $0.5$. 
Finally,
we use two stages for optimization to further improve the masks; after obtaining the initial 3D mask, the masks are rendered from the training views, and are used to supervise a secondary multiview segmentation model as initial guesses (instead of the video segmentation outputs).

\subsection{Multiview Inpainting}\label{sec:mv.inpaint}

\begin{figure}[t]
  \centering
   \includegraphics[width=0.99\linewidth]{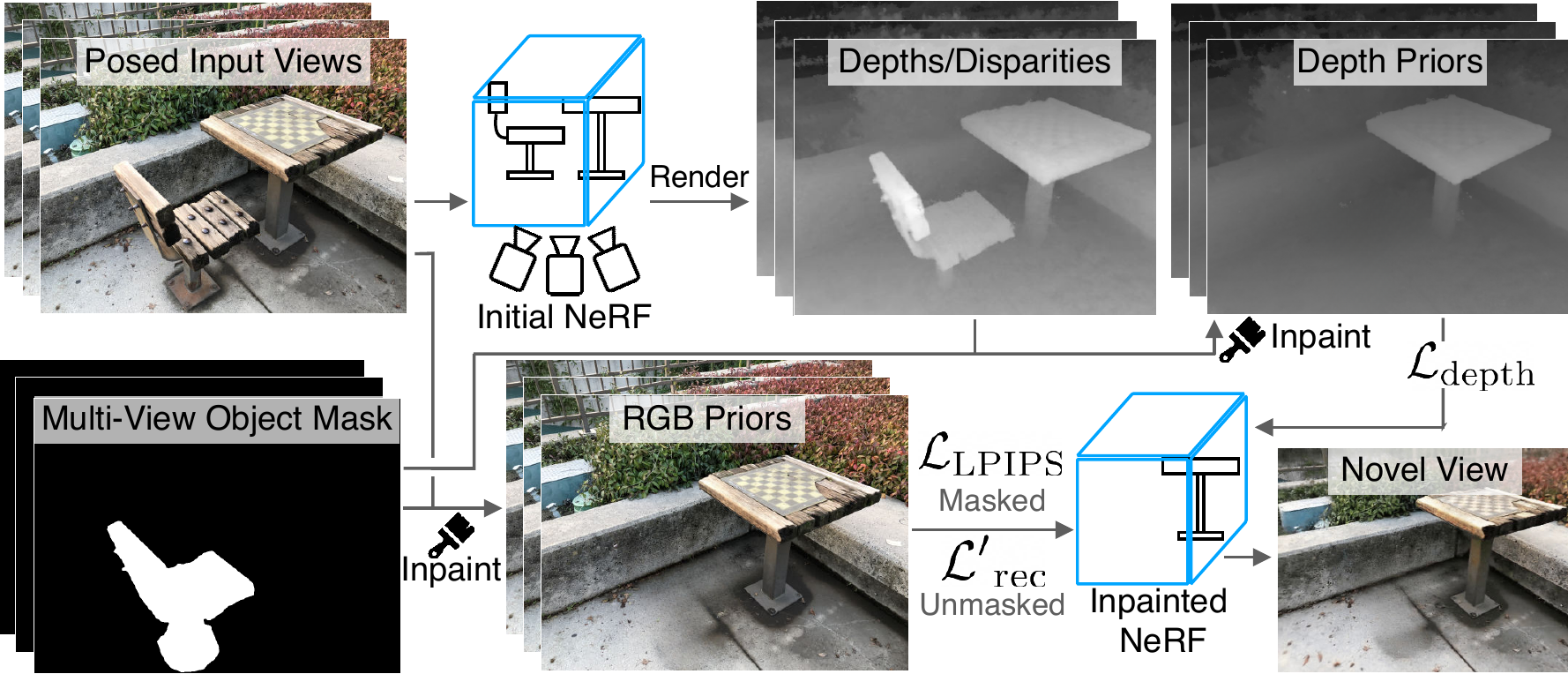}

   \caption[Caption for LOF]{
   Overview of our inpainting method. Using posed input images and their corresponding masks (upper- and lower-left insets), we obtain (i) an initial NeRF \textit{with} the target object present and (ii) the set of inpainted input RGB images with the target object removed (but with view inconsistencies). The initial NeRF (i) is used to compute depth, which we inpaint to obtain depth images as geometric priors (upper-right inset). The inpainted RGB images (ii), which act as appearance priors, are used with the depth priors, to fit a 3D consistent NeRF to the inpainted scene.\protect\footnotemark
   }
   \label{fig:inpainting.overview}
\end{figure}

\footnotetext{IBRNet images in Fig. 3,5,6,7 by Wang et al. available in IBRNet~\cite{ibrnet} under an \href{https://github.com/googleinterns/IBRNet/blob/master/LICENSE}{Apache License 2.0}.}

Figure~\ref{fig:inpainting.overview} shows an overview of our view-consistent inpainting method. 
As the paucity of data precludes directly training a 3D inpainter, our method leverages existing 2D inpainters to obtain depth and appearance priors, which then supervise the fitting of a NeRF to the completed scene.
This inpainted NeRF is trained using the following loss:
\begin{equation}
    \label{eq:full.loss}
    \mathcal{L}_\text{inp} = \mathcal{L'}_\text{rec} + \lambda_\text{LPIPS} \mathcal{L}_\text{LPIPS} + \lambda_\text{depth} \mathcal{L}_\text{depth},
\end{equation}
where $\mathcal{L'}_\text{rec}$ is the reconstruction loss for the unmasked pixels, and $\mathcal{L}_\text{LPIPS}$ and $\mathcal{L}_\text{depth}$
define the perceptual and depth losses (see below), with weights $\lambda_\text{LPIPS}$ and $\lambda_\text{depth}$.

\subsubsection{RGB Priors}

Our proposed view-consistent inpainting approach uses RGB inputs, $\{ I_i \}_{i=1}^{n}$, the camera intrinsic and extrinsic parameters, and corresponding object masks, $\{ M_i \}_{i=1}^{n}$, to fit a NeRF to the scene without the undesired object. To begin with, each image and mask pair, $(I_i, M_i)$, is given to an image inpainter, $\text{INP}$, to obtain the inpainted RGB images, $\{ \tilde I_i \}_{i=1}^{n}$, where $\tilde I_i = \text{INP}(I_i, M_i)$~\cite{lama}. Since each view is inpainted independently, directly supervising a NeRF using the inpainted views leads to blurry results due to the 3D inconsistencies between each $\tilde I_i$ (see Figure~\ref{fig:nerfin.comparison}). 
In this paper, instead of using mean squared error (MSE) to optimize the masked area, we propose the use of a perceptual loss, LPIPS~\cite{perceptual}, to optimize the masked parts of the images, while still using MSE for the unmasked parts, where no inpainting is needed. This loss is calculated as follows:
\begin{equation}
    \label{eq:perceptual.loss}
    \mathcal{L}_\text{LPIPS} = \frac{1}{\vert \mathcal{B} \vert} \sum_{i \in \mathcal{B}} \text{LPIPS}(\hat I_i, \tilde I_i),
\end{equation}
where $\mathcal{B}$ is a batch of indices between $1$ and $n$, 
and $\hat I_i$ is the $i$-th view rendered using NeRF. 
Our model for multiview inpainting and segmentation uses the same architecture (see Figure~\ref{fig:mv.seg.architecture}), 
except for the additional logit output, $s$.

\subsubsection{Depth Priors}

Even with the perceptual loss, the discrepancies between the inpainted views can incorrectly guide the model towards converging to degenerate geometries 
(e.g., `foggy' geometry may form near the cameras, to explain the disparate per-view information).
Thus, we use inpainted depth maps as additional guidance for the NeRF model, and detach the weights when calculating the perceptual loss and use the perceptual loss to only fit the colors of the scene. For this purpose, we use a NeRF optimized on images that include the unwanted object, and render the depth maps, $\{D_i\}_{i=1}^{n}$, corresponding to the training views. Depth maps are calculated by substituting the distance to the camera instead of the color of points into Eq.~\ref{eq:volumetric.rendering.discrete}:
\begin{equation}
    \label{eq:depth.volumetric.rendering.discrete}
    D(r) = \sum_{i=1}^{N} T_i (1 - \exp(-\sigma_i \delta_i)) t_i.
\end{equation}
The rendered depths are then given to an inpainter to obtain inpainted depth maps, $\{ \tilde D_i \}_{i=1}^{n}$, where $\tilde D_i$ is obtained as $\tilde D_i = \text{INP}(D_i, M_i)$. We found that using LaMa~\cite{lama} for depth inpainting, as in the RGB case, gave sufficiently high-quality results. Note that this is all calculated as a preprocessing step, and with a NeRF optimized on the original scene. This NeRF can be the same model used for multiview segmentation. If using another source for obtaining masks, such as human annotated masks, a new NeRF is fitted to the scene. 
These depth maps are then used to supervise the inpainted NeRF's geometry, via the $\ell_2$ distance of its rendered depths, $\hat D_i$, to the inpainted depths, $\tilde D_i$:
\begin{equation}
    \label{eq:depth.loss}
    \mathcal{L}_\text{depth} = \frac{1}{\vert \mathcal{R}\vert} \sum_{r \in \mathcal{R}} 
    \left| \hat D(r) - \tilde D(r) \right|^2,
\end{equation}
where $\hat D(r)$ and $\tilde D(r)$ are the depth values for a ray, $r$.

\subsubsection{Patch-based Optimization}

Calculating the perceptual loss, Eq.~\ref{eq:perceptual.loss}, requires full input views to be rendered during the optimization. Since rendering each pixel necessitates multiple forward passes through the MLP, for high-resolution images, this is an expensive process, resulting in issues such as
\begin{enumerate*}[label=(\roman*)]
    \item the batch size, $\vert \mathcal{B} \vert$, has to be small to fit the rendered images and their corresponding computation graphs in memory, and 
    \item slow optimization, even with batch sizes as small as $\vert \mathcal{B} \vert = 1$. 
\end{enumerate*}
A straightforward solution is to render a downsized image and compare it to the downsized version of the inpainted images; however, this leads to a loss of information if the downsizing factor is large. Following image-based works (e.g., SinGAN~\cite{singan} and DPNN~\cite{drop.the.gan}), and 3D works (e.g., ARF~\cite{arf}), we perform the computations on a patch-basis; instead of rendering complete views, we render batches of smaller patches, and compare them with their counterparts in the inpainted images based on the perceptual loss. 
Only patches inside the bounding box of the object mask are used. 
For fitting the unmasked areas, recall that $\mathcal{L'}_\text{rec}$ (Eq.~\ref{eq:full.loss}) simply alters $\mathcal{L}_\text{rec}$ (Eq.~\ref{eq:reconstruction.loss}) to sample rays only from unmasked pixels.
By separating the perceptual and reconstruction losses, 
we prevent inconsistency within the mask, while avoiding unnecessary changes to the rest of the scene.

\subsubsection{Mask Refinement}\label{sec:mask.refinement}

Here, we consider further leveraging the multiview data to guide the image inpainter. In particular, parts of the training images that are currently being generated by the 2D image inpainter might be visible in other views; in such cases, there is no need to hallucinate those details, since they can be retrieved from the other views. To prevent such unnecessary inpaintings, we propose a mask refinement approach: for each source image, depth, and mask tuple, $(I_s, D_s, M_s)$, we substitute pixels in $I_s$ and $D_s$ that are visible from at least one other view, to shrink the source mask, $M_s$. After this refinement step, only parts of $I_s$ and $D_s$ that are occluded by the undesired object in \textit{all} of the training views will remain masked. As a result, the image inpainter has to fill in a smaller area, resulting in improved inpaintings. 
Please see our supplementary material for details.

\section{Experiments}
\label{sec:experiments}

\begin{figure}[t]
  \centering
   \includegraphics[width=0.99\linewidth]{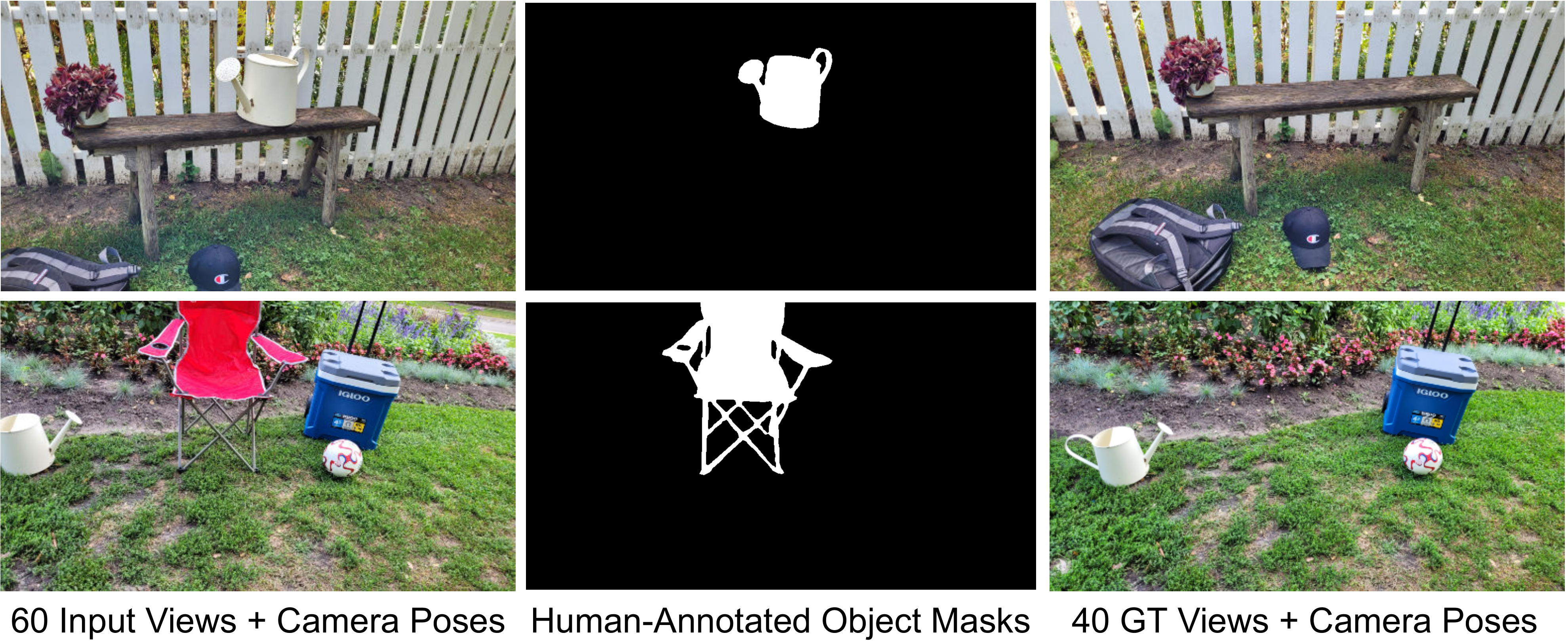}

   \caption{
    Scenes from our dataset.
    Columns: input view (left), corresponding target object mask (middle), and a ground-truth view without the target object, from a different camera pose (right).
    Rows: different scenes; see supplement for examples of all scenes. %
   }
   \label{fig:dataset.samples}
\end{figure}

\textbf{Dataset}.
To evaluate multiview segmentation (MVSeg), we adopt real-world scenes from LLFF~\cite{llff}, NeRF-360~\cite{original.nerf}, NeRF-Supervision~\cite{nerf.supervision}, and Shiny~\cite{nex}. For multiview (MV) inpainting, in addition to providing qualitative results on scenes from IBRNet~\cite{ibrnet}, we address the need for a standard benchmark including ground-truth captures of scenes without the unwanted object as test views, and introduce a dataset containing 10 real-world forward-facing scenes with human annotated object masks. For each scene, we provide 60 training images with the object and 40 test images without the object. This dataset is further suitable for evaluating tasks such as real-time 3D inpainting, unsupervised 3D segmentation, and video inpainting. Figure~\ref{fig:dataset.samples} shows sample views from two scenes of our dataset. 

\begin{table}[tb]
\centering
\caption{Quantitative evaluation of multiview segmentation models, for the task of transferring the source mask to other views. }
\resizebox{0.48\textwidth}{!}{
\begin{tabular}{lcc}
\hline
\multicolumn{1}{l}{\textbf{Method}} & \textbf{Acc.}$\uparrow$ & \textbf{IoU}$\uparrow$ \\ \hline
Proj.\ + Grab Cut\cite{grab.cut} (2D)                                 & 91.08                         & 46.61              \\
Proj.\ + EdgeFlow\cite{edge.flow} (2D)                                & 96.84                         & 81.63              \\ \hdashline
Semantic NeRF~\cite{semantic.nerf} (only source mask)                  & 94.63                         & 75.13              \\
Proj.\ + EdgeFlow\cite{edge.flow} + Semantic NeRF\cite{semantic.nerf}  & 97.26                         & 83.95              \\
Feature Field Distillation~\cite{feature.fields}                      & 97.37                         & 83.07              \\
Video Segmentation\cite{video.seg.survey,dino}                        & 98.43                         & 88.34              \\
Ours                                                                  & 98.85                         & 90.96              \\
Ours (two-stage)                                                      & \textbf{98.91}                & \textbf{91.66}  \\ \hline      
\end{tabular}
\label{tab:multiview.segmentation}
}
\end{table}

\textbf{Metrics}.
To evaluate our segmentation model, we use the accuracy of the predictions (pixel-wise) and the intersection over union (IoU) metric. For MV inpainting, we follow the image-to-image literature~\cite{lama} and report the average learned perceptual image patch similarity (LPIPS)~\cite{perceptual}, and the average Fréchet inception distance (FID) \cite{fid} between the distribution of the ground-truth test views and model outputs. Since our focus is the inpainting, we only calculate the LPIPS and FID inside the bounding box of the object mask (using our MVSeg model, we can render the object mask from the test views that do not contain the object).

\textbf{Multiview Segmentation Baselines}.
For MVSeg, one category of baselines is projection-based approaches: the source mask is projected into the other views using the scene geometry from a NeRF. This gives us an incomplete mask in the other views. Then, a variety of interactive segmentation approaches are applied to the incomplete masks, propagating them to obtain complete object masks: \textit{Proj.\ + Grab Cut~\cite{grab.cut}} and \textit{ Proj.\ + EdgeFlow~\cite{edge.flow}}. In addition, we consider \textit{Proj.\ + EdgeFlow + Semantic NeRF}, where an additional Semantic NeRF~\cite{semantic.nerf} is fitted to make the outputs 3D consistent. Another baseline~\cite{feature.fields} is a representative of the concurrent works on distilling 2D pixel-level features to 3D scenes~\cite{feature.fields.concurrent} and post-processing them for obtaining segmentation masks. As a baseline for video-segmentation~\cite{video.seg.survey}, we compare to Dino~\cite{dino} since it does not rely on temporally close neighboring frames. 

\textbf{Multiview Inpainting Baselines}.
Masked NeRF only uses the unmasked pixels to optimize a NeRF. Object NeRF~\cite{yang2021learning} filters the unwanted points in 3D without explicitly inpainting the missing regions. No code is available for NeRF-In~\cite{nerf.in}, so we use our own implementation of their model, with slight modifications (we use LaMa\cite{lama} as the inpainter). In addition, we compare our results to LaMa~\cite{lama} as a representative of state-of-the-art 2D inpainters. To enable fair comparison between the NeRF-based 3D models (which, in addition to inpainting, have to synthesize novel views), we compare to LaMa by (i) fitting a NeRF on the views \textit{with} the object\footnote{Since the test views that do not contain the object should not be available to the model during the inference.}, (ii) rendering the test views from the fitted NeRF, and (iii) passing these rendered images to LaMa.
Finally, for reference and as an ideal ``gold standard'' 3D inpainting baseline, we fit a NeRF on the \textit{ground-truth} test images, use the optimized NeRF to render the test-views, and then compare the rendered results to the ground-truth. We provide these results for completeness, as an upper-bound on the best possible results one can expect to obtain when using the same NeRF architecture. 

\begin{figure*}[t]
  \centering
   \includegraphics[width=0.99\linewidth]{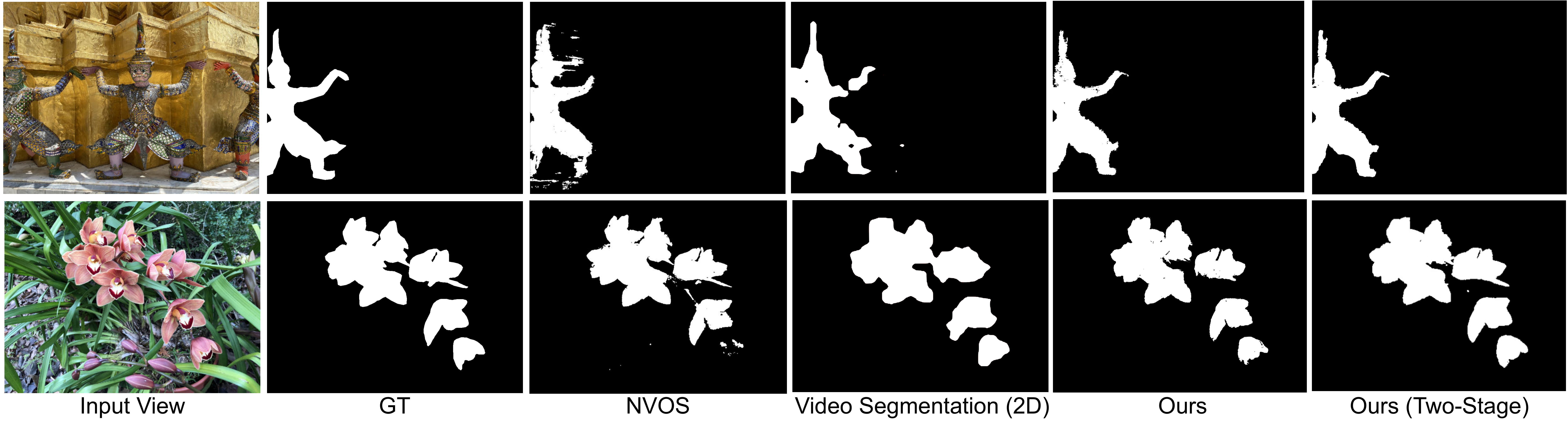}
   \caption{
      A qualitative comparison of our multiview segmentation model against Neural Volumetric Object Selection (NVOS)~\cite{neural.object.selection} (reproduced from original paper), video segmentation~\cite{dino}, and the human-annotated masks (GT).
      Our two-stage approach refers to running our multiview segmentation model twice, with the output of the first run as the initialization of the second
      (see \autoref{sec:mv.seg.nerf}).
      Our method is less noisy than NVOS, which also misses some pieces of the target (e.g., the lowest flower in the bottom row), but captures more details than the video segmentation alone (e.g., the blurred edges of the flowers). 
      Our two-stage approach helps fill in some missing aspects of the one-stage output. Please see our supplement for additional qualitative examples. 
   }
   \label{fig:mv.seg.qualitative}
\end{figure*}

\subsection{Results}

\textbf{Multiview Segmentation}.
We first evaluate our MVSeg model without any inpainting. In this experiment, we assume that the sparse image points are already given to an off-the-shelf interactive segmentation model, and that the source mask is available. Thus, the task is to transfer the source mask to other views.  Table~\ref{tab:multiview.segmentation} shows that our model outperforms all of the 2D (3D-inconsistent) and 3D-consistent baselines. In addition, our two-stage optimization helps to further improve the obtained masks.

Qualitatively, Figure~\ref{fig:mv.seg.qualitative} compares the results of our segmentation model with the outputs of Neural Volumetric Object Selection (NVOS)~\cite{neural.object.selection} and video segmentation~\cite{dino, video.seg.survey}. Compared to the coarse edges of the 3D-unaware video segmentation model, our model reduces noise and improves consistency across views. Although NVOS uses scribbles, rather than the sparse points used in our model, our MVSeg visually outperforms NVOS. Since the NVOS code base is not available,
we reproduce the published qualitative results for NVOS~\cite{neural.object.selection} 
(see supplemental for more examples). 

\begin{table}[tb]
\centering
\caption{Quantitative evaluation of our inpainting method, using human-annotated object masks. }
\begin{tabular}{lcc}
\hline
\multicolumn{1}{l}{\textbf{Method}} & \textbf{LPIPS}$\downarrow$ & \textbf{FID}$\downarrow$\\ \hline
Ideal                                & 0.4079                         & 100.25\\ \hdashline
LaMa (2D)~\cite{lama}                & 0.5369                         & 174.61 \\
Object NeRF~\cite{yang2021learning}  & 0.6829                         & 271.80\\
Masked NeRF~\cite{original.nerf}     & 0.6030                         & 294.69\\
NeRF-In~\cite{nerf.in}               & 0.5699                         & 238.33\\
NeRF-In~\cite{nerf.in} (Single)      & 0.4884                         & 183.23\\
Ours (no geo. inpainting)            & 0.4952                         & 200.34\\
Ours                                 & 0.4654                         & 156.64\\
Ours (Refined RGB/Depth)             & 0.4664                         & 163.79\\
Ours (Refined RGB)                   & \textbf{0.4529}                & \textbf{147.31}\\\hline
\end{tabular}
\label{tab:multiview.inpainting}
\end{table}

\textbf{Multiview Inpainting}.
Our dataset is used for quantitative evaluation of our proposed inpainting method against the baselines. 
Table~\ref{tab:multiview.inpainting} shows the comparison of our MV inpainting method with the baselines, assuming that the object masks from all of the views are given. The ``ideal'' row is not a baseline, but rather a NeRF fitted to the ground-truth test views (views of the scene without the object). 
While this is only one instantiation of the many possible inpainted scenes, it provides a convenient measure of the best performance one might reasonably expect in this scenario.
Overall, our method significantly outperforms the alternative 2D and 3D inpainting approaches. 
Although our model uses a 2D image inpainter's~\cite{lama} outputs to obtain a view-consistent inpainted NeRF, it is able to use this ensemble of MV information, in addition to the priors encoded by the perceptual loss function, to outperform the 2D inpainter. 
Table~\ref{tab:multiview.inpainting} further shows that removing geometry guidance impairs the inpainted scene quality. 

\begin{table}[tb]
\centering
\caption{Quantitative evaluation of our inpainting method, using the outputs of our multiview segmentation model. 
}
\begin{tabular}{lcc}
\hline
\multicolumn{1}{l}{\textbf{Method}} & \textbf{LPIPS}$\downarrow$ & \textbf{FID}$\downarrow$\\ \hline
LaMa (2D)~\cite{lama}                & 0.5439                         & 169.92\\
Object NeRF~\cite{yang2021learning}  & 0.6679                         & 286.55\\
Masked NeRF~\cite{original.nerf}     & 0.6340                         & 332.70\\
NeRF-In~\cite{nerf.in}               & 0.5858                         & 240.27\\
NeRF-In~\cite{nerf.in} (Single)      & 0.5054                         & 176.27\\
Ours                                 & \textbf{0.4662}                & \textbf{140.56}\\ \hline
\end{tabular}
\label{tab:multiview.inpainting.mvseg}
\end{table}

\begin{figure*}[t]
  \centering
  \includegraphics[width=0.99\linewidth]{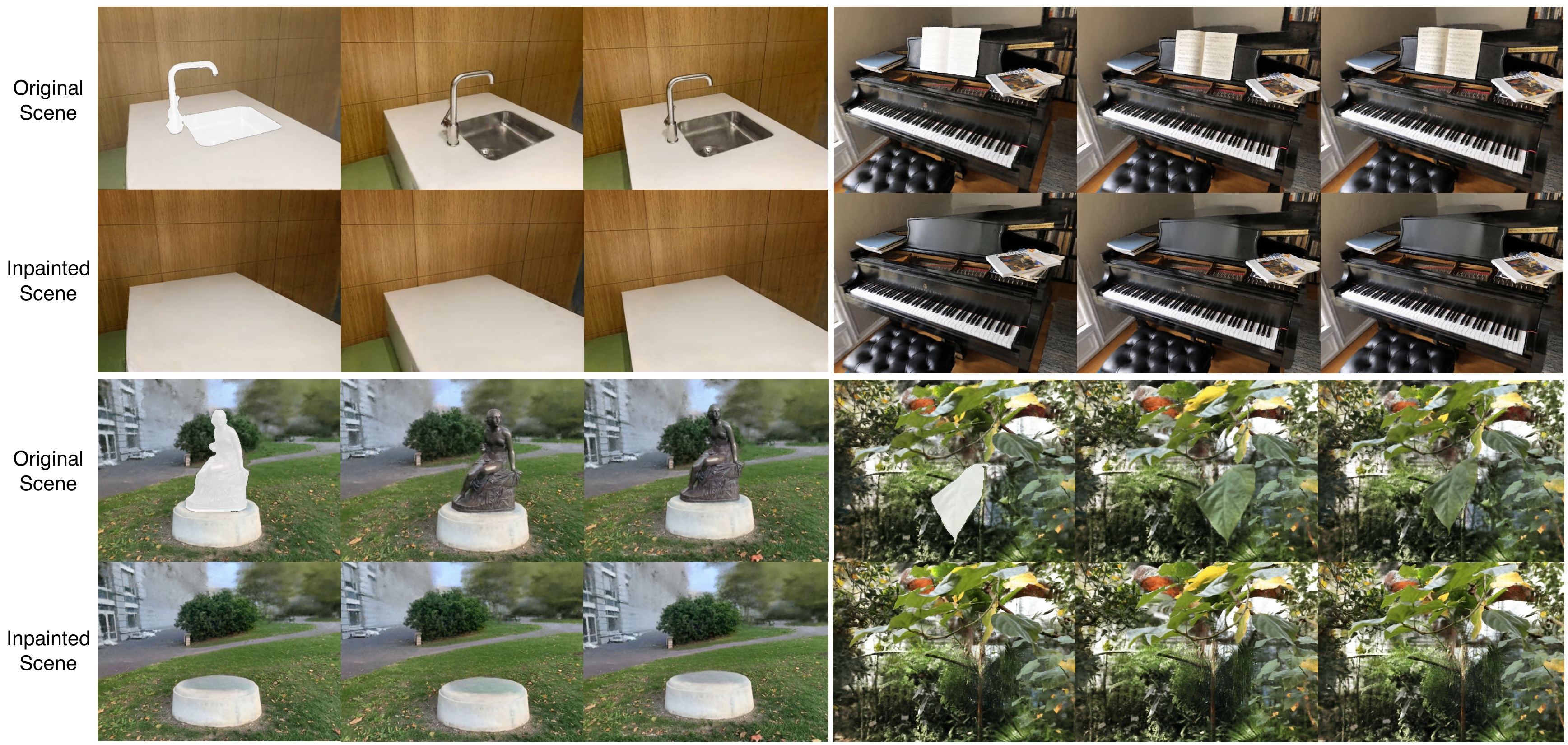}
  \caption{
    Visualizations of our view-consistent inpainting results.
    Upper rows per inset show NeRF renderings of the original scene from novel views, with the first image also displaying the associated mask.
    Lower rows show the corresponding inpainted view.
    Notice that the synthesized views maintain consistency with each other;
    however, view-dependent effects still remain (e.g., the lighting on the uncovered part of the piano). Please see our supplement for additional scenes, and our \href{https://spinnerf3d.github.io}{project website} with videos for better visualization. 
  }
  \label{fig:inpainting.qualitative}
\end{figure*}

\begin{figure}[t]
  \centering
  \includegraphics[width=0.99\linewidth]{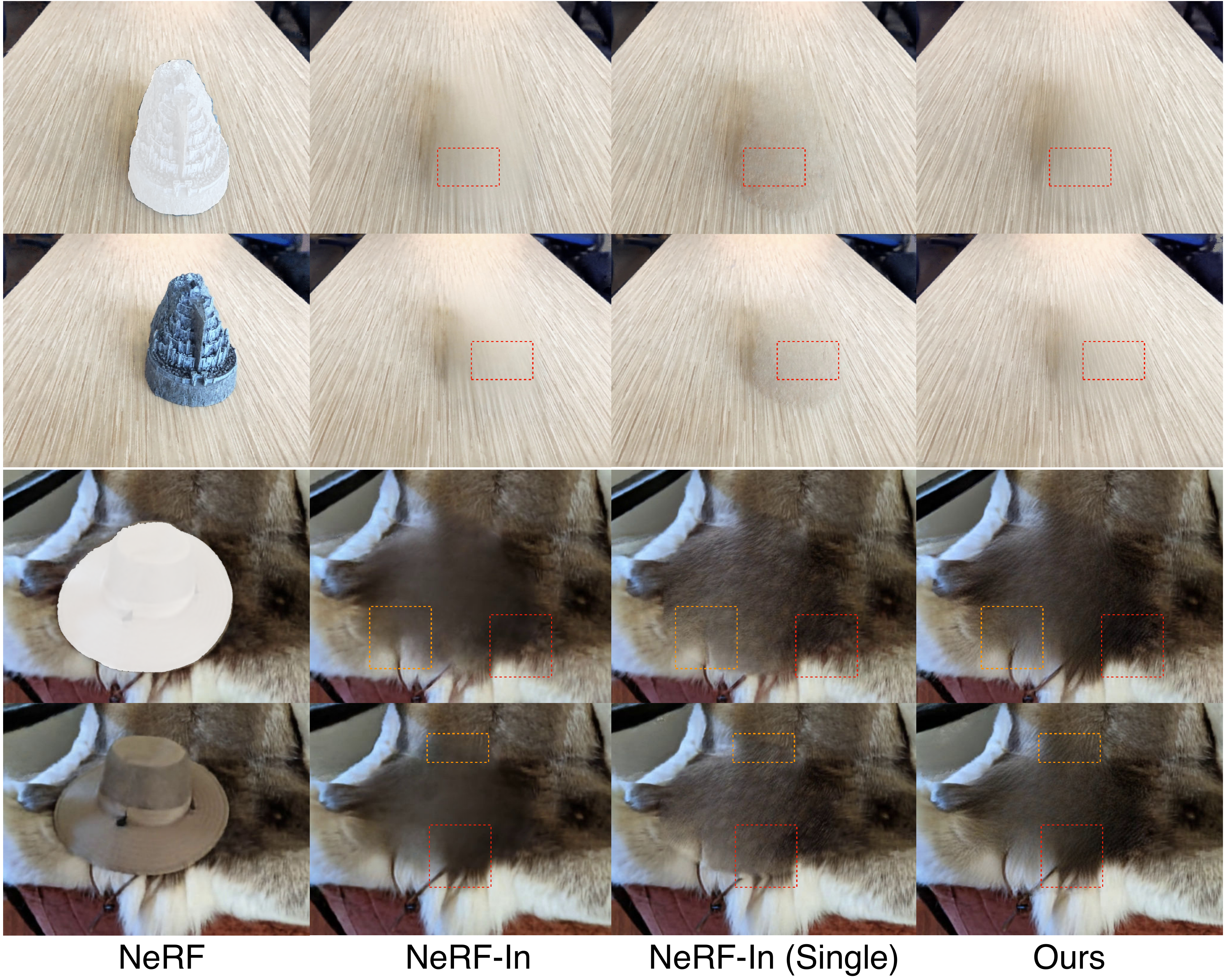}
  \caption{
    Qualitative comparisons with other baselines.
    Columns: novel views of the scene, synthesized by a NeRF (on the unmasked images), NeRF-In, NeRF-In (with a single masked training image), and our approach.
    NeRF-In is significantly more blurry, while NeRF-In (single) tends to have difficulty with details closer to the edge of the mask boundary (zoom into boxes for examples).
  }
  \label{fig:nerfin.comparison}
\end{figure}

We display qualitative results of our MV inpainting method in Figure~\ref{fig:inpainting.qualitative}, showing that it can reconstruct a view-consistent scene with detailed textures, including coherent view-dependent radiance for both shiny and non-shiny surfaces. 
In addition, in Figure~\ref{fig:nerfin.comparison}, we provide a visual comparison to the unpublished concurrent work NeRF-In~\cite{nerf.in}, 
which has the second-lowest error.
We observe that a NeRF-In model fitted to all of the inpainted views results in blurry outputs. Alternatively, using a single inpainted view for supervising the masked region leads to artifacts in further views, due to the lack of supervision for the view-dependent radiance, as well as the poor extrapolation capabilities of the network.
In contrast, our perceptual approach relaxes the exact reconstruction constraints in the masked region, thus preventing blurriness despite using all of the images, while avoiding artifacts caused by single-view supervision.

Table~\ref{tab:multiview.inpainting} shows that refinement provides a small but significant boost in inpainting quality, due to smaller masks requiring less hallucination from the inpainter. Yet, empirically, 
refinement only subtly trims the masks
(reducing mean masked area by $4.74$\% on our dataset), as the camera has limited movement during data collection, to ensure similarity between the training and testing views.
Further, due to noisy NeRF geometries projecting incorrect values, refining \textit{depths} lowers performance, whereas refining \textit{colours alone} achieves our best results.

So far, our experiments examine the performance of our MVSeg and MV inpainting independently; however, one can combine them to remove objects from NeRF scenes with minimal user interaction. 
Table~\ref{tab:multiview.inpainting.mvseg} shows that using the output masks of our MVSeg model, instead of using the human-annotated object masks, results in a subtle decrease in the inpainting quality.
However, our model with MVSegs still outperforms other methods, even when they are fitted on \textit{human}-annotated segmentations.

\begin{table}[tb]
\centering
\caption{Evaluation of our 3D inpainting method across input view number (upper half) and mask dilation level (lower half). } 
\begin{tabular}{lcccc}
\hline
\multicolumn{1}{l}{\textbf{Input Views}} & \textbf{10} & \textbf{20} & \textbf{40} & \textbf{60} \\ \hline
\textbf{LPIPS}$\downarrow$ & 0.4726 & 0.4713 & 0.4667 & \textbf{0.4654}  \\
\textbf{FID}$\downarrow$   & 171.19 & 172.63 & 158.02 & \textbf{156.64} \\ 
\hline %
\multicolumn{1}{l}{\textbf{Dilation}} & \textbf{45} & \textbf{25} & \textbf{5} & \textbf{0} \\ \hline
\textbf{LPIPS}$\downarrow$ & 0.6102 & 0.5369 & \textbf{0.4654}  & 0.4904   \\
\textbf{FID}$\downarrow$   & 283.01 & 230.03 & \textbf{156.64} & 164.66 \\  \hline
\end{tabular}
\label{tab:mask.ninputs.dilation}
\end{table}

\subsection{Variations and Ablation Studies}

\textbf{Number of Input Views}.
Limiting the number of input views is a standard approach employed in the literature to modulate the reconstruction quality of NeRFs~\cite{nesf}. Table~\ref{tab:mask.ninputs.dilation} shows that the performance of our inpainter degrades with fewer inputs. Thus, we argue that as better-quality NeRFs are introduced, our approach, which is agnostic to the underlying NeRF model, can readily benefit.

\textbf{Importance of Accurate Masks}.
Here, we examine the impact of accurate masks on inpainting, via variable dilations of the object masks with a $5\times5$ kernel.
Larger masks lead to relying more on the view-inconsistent outputs and hallucinations of the 2D inpainters, while smaller masks may permit parts of the unwanted object's edges to remain and confuse the 2D inpainter.
A subtle dilation is also useful for reducing the effect of shadows.
This balance between over-masking and under-masking is demonstrated in Table~\ref{tab:mask.ninputs.dilation}, with five dilation iterations found to be optimal and therefore used for all other experiments.

\section{Conclusion}
\label{sec:conclusion}

In this paper, we presented a novel approach to inpaint NeRF scenes, which enforces viewpoint consistency based on image and geometric priors, given a single-view object mask. In addition, we provided a multiview segmentation method that simplifies the annotation process by using a set of sparse pixel-level clicks on (and around) the undesired object and translating them into a 3D mask that can be rendered from novel views. We provided experiments to show the effectiveness of our segmentation and inpainting methods. 
The main limitation of our work is the assumption of semantically consistent image priors, potentially only differing in terms of textures. 
Finally, we introduce a dataset that not only addresses the lack of challenging benchmarks for multiview inpainting, but which we believe can assist future advances in this new line of research.

\textbf{Acknowledgments}.
This work was conducted at Samsung AI Centre Toronto and it was funded by Mitacs and Samsung Research, Samsung Electronics Co., Ltd.

{\small
\bibliographystyle{ieee_fullname}
\bibliography{egbib}
}

\newpage
\appendix

\section{Summary}

We provide details of our proposed refinement process in~\autoref{sec:refinement.details}. In~\autoref{sec:additional.details}, additional details about our implementation are provided for reproducibility. \autoref{sec:additional.qualitative} contains further qualitative experiments showing the visual performance of our multiview segmentation and inpainting methods. We also provide a supplementary video and a website with video renderings of the scenes with and without inpainting for better visualization. In~\autoref{sec:mv.seg.multistage}, we provide an ablation study measuring the impact of additional training stages to segmentation performance. \autoref{sec:mv.dataset.appendix} provides an overview of all of the scenes in our introduced dataset. For completeness, we provide an extended version of the background on NeRFs in~\autoref{sec:extended.bg}. A detailed version of the segmentation results can be found in~\autoref{detailed.segmentation.results}. In~\autoref{failure.cases}, we discuss potential failure cases of our model. Finally, due to the generative nature of inpainting, we provide an ethics statement in~\autoref{sec:ethical.statement}.

\section{Refinement Details}
\label{sec:refinement.details}

For pixel values that are only visible in some of the views, we use mask refinement to project them to all of the input views, as introduced in~\autoref{sec:mask.refinement} in the main paper. This refinement reduces the masked area and leads to better inpaintings due to a decreased need for hallucination. Consider a source image, $I_s$, its corresponding depth, $D_s$, and mask, $M_s$. For each target image, depth, and mask tuple, $(I_t, D_t, M_t)$, and for every masked pixel in the source view, $u_s$, we consider the ray passing through $u_s$: $r_{u_s} = o_{u_s} + t d_{u_s}$. The same sampling approach used in the original NeRF paper~\cite{original.nerf} is performed to sample $\{t_i\}_{i=1}^{N}$ on ray $r_{u_s}$. At the $i$-th step, the point represented by $t_i$ is projected into the world coordinate system as:
\begin{equation}
    \label{eq:project.to.world}
    X_i = G_s K^{-1} t_i u_s, 
\end{equation}
where $G_s$ is the source camera pose and $K$ is the camera intrinsic matrix. Next, point $X_i$ is unprojected into the target views to determine which pixel in the target view corresponds to $u_s$~\cite{nerf.supervision}:
\begin{equation}
    \label{eq:unprojection}
    u_{t, i} = \pi (K G_t ^{-1} X_i), 
\end{equation}
where $G_t$ is the camera pose of the target view, and $\pi$ stands for the perspective projection operation. If $u_{t, i}$ is masked in the target view, $t_i$ is ignored and we go to $t_{i + 1}$. If it is not masked, we check if the depth, $D_t(u_{t, i})$, is consistent with the distance of $X_i$ to the target camera. In case of depth inconsistency, again, $t_i$ is discarded and we proceed to $t_{i + 1}$. If the depths are consistent, the RGB color $I_s(u_s)$ is replaced with $I_t(u_{t, i})$ while unmasking $u_s$ in the source view. Note that for refining $D_s(u_s)$, one cannot directly use $D_t(u_{t, i})$ because it is the distance to the source camera. The depth $D_t(u_{t, i})$ is first projected to the world coordinates similar to \ta{Kosta suggests a figure, maybe for the suppmat?} Eq.~\ref{eq:project.to.world} as:
\begin{equation}
    \label{eq:depth.projection.to.world}
    X_\text{depth} = G_t K^{-1} D_s(u_{t, i}) u_{t, i}.
\end{equation}
The distance of $X_\text{depth}$ to the source camera is then used to replace $D_s(u_s)$.

\begin{figure}[t]
  \centering
   \includegraphics[width=1.0\linewidth]{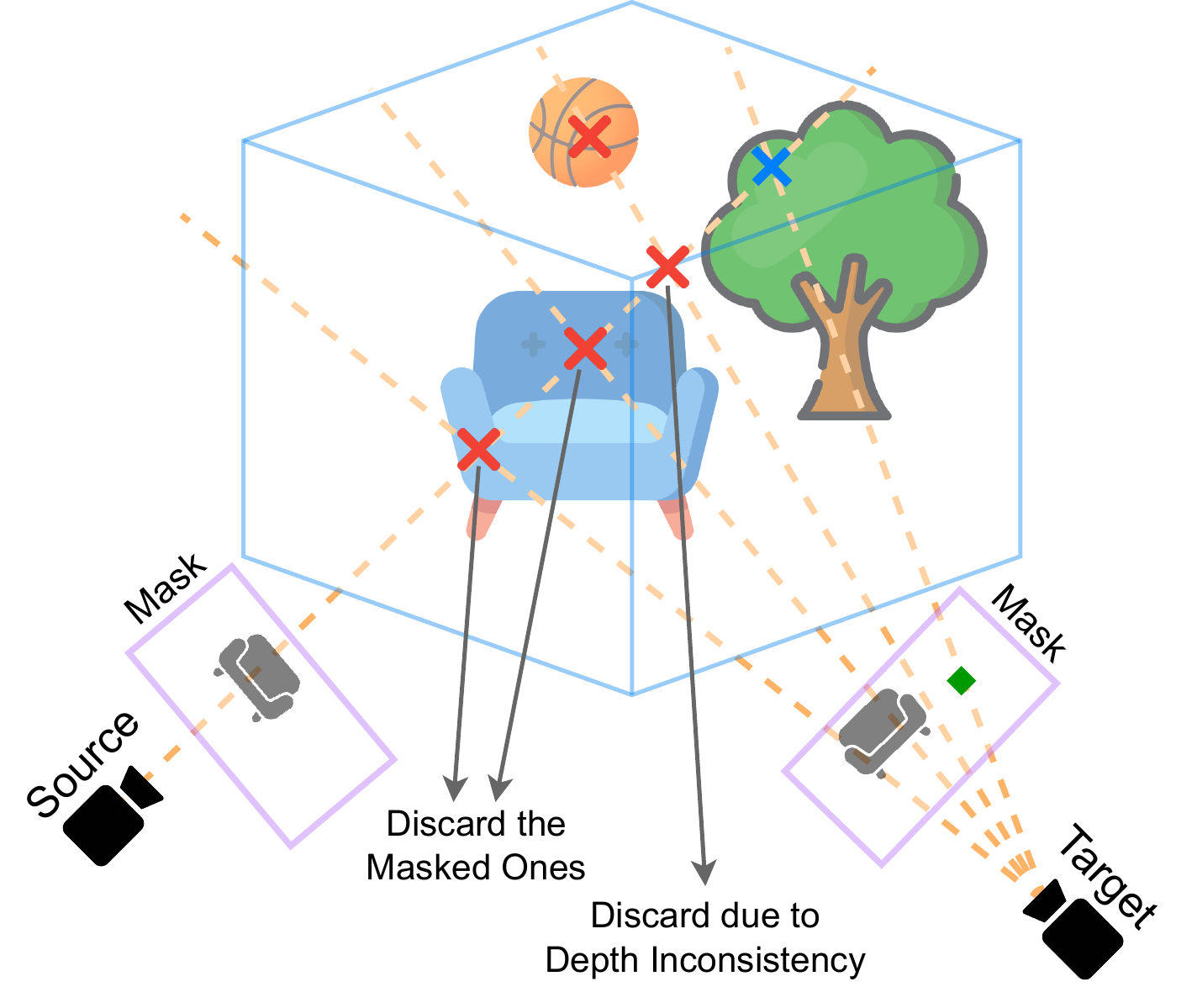}

   \caption{A visualization of our proposed mask refinement. The green pixel depicted in the target view is the final one that is used to transfer color and depth from the target view to the source view. Crosses represent the sampled points, and the blue cross is the final point used for the refinement in this example. }
   \label{fig:mask.refinement}
\end{figure}

For each source image, we visit the target images one by one; if a pixel is able to be refined with respect to a new target image, the refinement is performed, and if a previously refined pixel is able to be refined with a point closer to the source camera, the refinement is updated. We iterate our refinement process multiple times, until no pixel is refined. This makes the process independent of the order of the target views. 

\begin{figure}[t]
  \centering
   \includegraphics[width=1.0\linewidth]{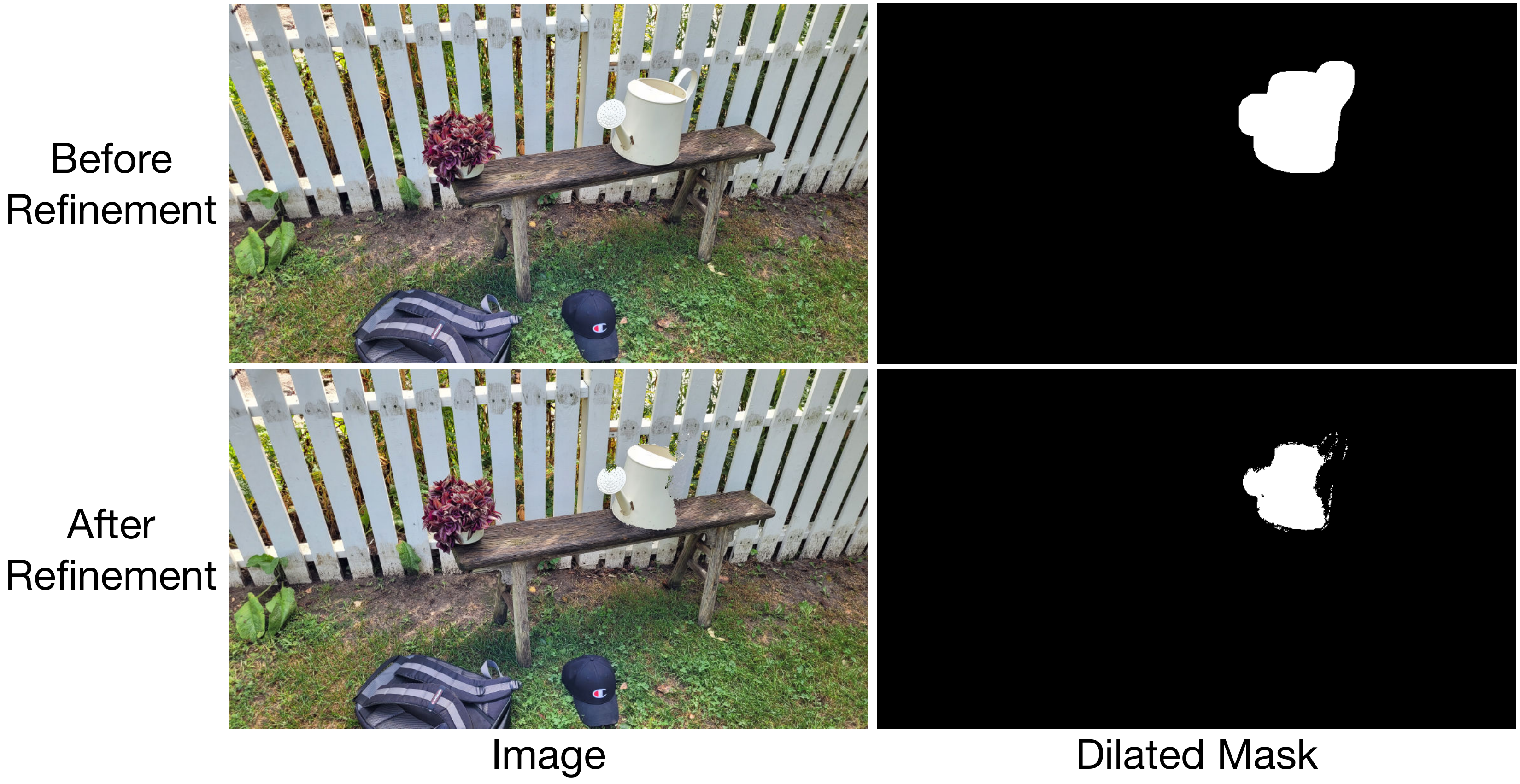}

   \caption{Qualitative example of how refinement can reduce the masked area by substituting pixel values from other views. }
   \label{fig:mask.refinement.qual}
\end{figure}

Figure~\ref{fig:mask.refinement} shows a toy example to visualize the mask refinement process. The unwanted object is the sofa. For a source and target view, a masked pixel in the source view is considered, and a ray is passed through this pixel. The first two sampled points on this ray are still masked when unprojected into the target view since they fall on the sofa in the 3D world. The next sampled point is unprojected to an unmasked pixel on the target view, but the depth is inconsistent since the target camera sees the basketball from that pixel. Finally, the blue cross shows the fourth sampled point, where the depth is consistent, and the green pixel corresponding to the leaves of the tree is used to refine the source image. The distance of the blue cross to the source camera is used to replace the source depth. In practice, a source pixel is refined only if, after the refinement, the new depth is consistent with at least one of the eight neighbouring pixels in the source view. Figure~\ref{fig:mask.refinement.qual} shows an example of an image from one of the scenes in our dataset, before and after refinement. We also provide corresponding masks to show the effect of our refinement process in reducing the masked area. Note that, following our other experiments in the main paper, the mask before refinement is dilated for five iterations, with a $5 \times 5$ kernel.

\section{Additional Details}
\label{sec:additional.details}

In practice, $\lambda_\text{LPIPS}$ and $\lambda_\text{depth}$ are set to $0.01$ and $1$, respectively. Our implementation is primarily in PyTorch~\cite{pytorch}, except for the encoders and MLP implementation, which use Tiny Cuda NN~\cite{tiny.cuda.nn} for efficiency. The models are trained on a single Nvidia RTX A6000 GPU. We use the sparse depth supervision in the unmasked regions of the input views, as in DS-NeRF~\cite{ds.nerf}, to obtain more accurate scene geometries. \ta{cite colmap as the source of the supervision?}
Following Instant-NGP~\cite{instant.ngp,hash.nerf}, the multi-resolution hash encoder used in our NeRF has $16$ levels, each returning two features. The base resolution is set to $16$. The MLPs have $64$-dimensional hidden layers. The first MLP, which calculates the density, $\sigma$ (and ``Objectness logit", $s$, for multiview segmentation), has two layers, while the color MLP has three layers. The training images used for our quantitative experiments have $567\times1008$ pixels (after being downsized four times to avoid memory issues), and all are captured by a Samsung Galaxy S20 FE. To calculate the perceptual loss, at each iteration, a random batch of four views is selected, and for each of them, a patch is rendered and compared to its inpainted counterpart in the perceptual space. Each patch is $16$ times smaller than the original image in each direction, while the stride for sampling the patches is set to two to cover larger areas. This makes the perceptual loss more meaningful, without slowing down the training. As mentioned in the paper, FID and LPIPS are calculated only for the bounding box of the masked region. The mask for test views is rendered using our multiview segmentation model, because the test views do not contain the object and can not be manually masked. Since in the experiments, masks are sometimes dilated, we also expand each side of the bounding box containing the mask in every direction by $10\%$ to make sure that in all of the experiments, the entire hallucinated region is being evaluated. 
Note that, for NeRF fitting, the object masks are slightly dilated (for five iterations with a $5\times 5$ kernel) to reduce the effects of the shadow of the target objects in the inpainted scene and to make sure that the mask covers all of the object. 

\textbf{Dataset}.
All scenes in our dataset are forward-facing, and obtained by manually moving a camera using an unstructured trajectory mimicking the behavior of a non-expert user. We focus on forward-facing scenes due in part to the fact that the inpainting task is more challenging, due to a lower chance to see behind objects and thus a need for more hallucination compared to $360^{\circ}$ scenes. All the $60+40$ images are jointly processed with Colmap to recover the camera parameters in a shared coordinate system.  Each image is $2268 \times 4032$ pixels in size.

\subsection{Approximate Timings}
In Table~\ref{tab:approx.timings}, we provide the approximate times that each stage in our framework takes. We use a similar architecture to Instant-NGP~\cite{instant.ngp}, which yields fast convergence for our models. Note that the semantic NeRF typically converges to an acceptable geometry even half-way through the fitting iterations, and the remaining iterations are mostly for obtaining a sharp appearance. Since our segmentation and inpainting approaches only use the rendered masks and depths from the semantic NeRF, according to the application, one can trade off quality for speed, and early stop the semantic NeRF to further reduce the segmentation time. For fitting the inpainted NeRF, since we have to render multiple patches and calculate the perceptual loss for each of them, the entire process is slower than the segmentation part. However, according to the fitting times in the literature, this is still a fast NeRF manipulation model for realistic scenes. Note that all of these times can be reduced in the future with faster hardware and underlying models, e.g., better differentiable scene representations. 

\begin{table}[tb]
\centering
\caption{Approximate times that each of the stages in our multiview segmentation and multiview inpainting framework take. These numbers do not include the time spent for human-annotations. }
\begin{tabular}{ll}
\hline
\multicolumn{1}{l}{\textbf{Stage Name}} & \textbf{Time}\\ \hline
Multiview Segmentation          &\\ \hdashline
Interactive Segmentation        & $<1$ second\\
Video Segmentation              & $<1$ minute\\
Fitting the Semantic NeRF       & $2-5$ minutes\\
Rendering Training Masks        & $1$ minute\\ \hline
Multiview Inpainting            &\\ \hdashline
Applying the Image Inpainter    & $<1$ minute\\
Fitting the Inpainted NeRF      & $20-40$ minutes\\
\hline      
\end{tabular}
\label{tab:approx.timings}
\end{table}

\section{Additional Qualitative Results}\label{sec:additional.qualitative}
\begin{figure*}[t]
  \centering
  \includegraphics[width=1.0\linewidth]{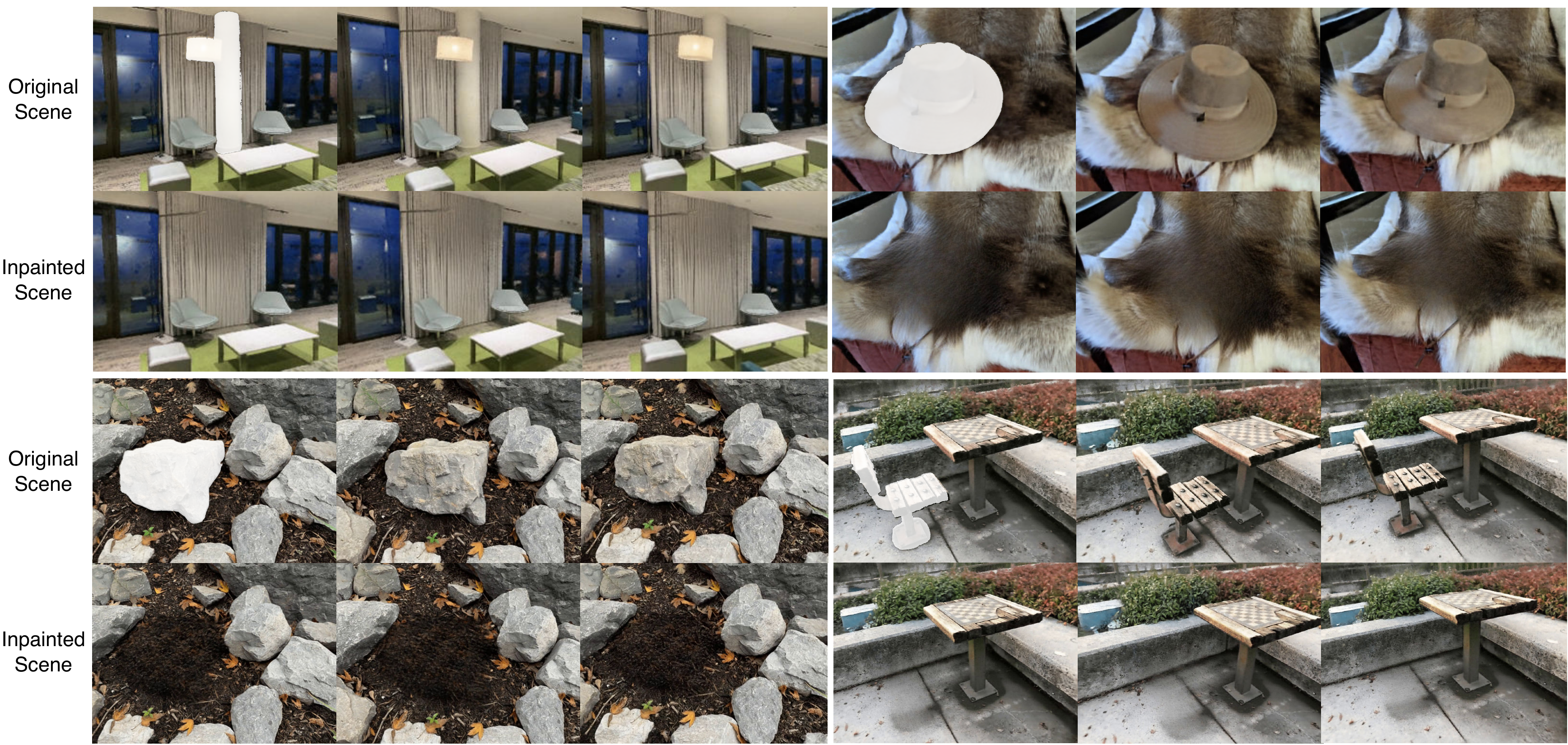}

  \caption{Additional qualitative visualizations of our view-consistent inpainting results, as in Figure~\ref{fig:inpainting.qualitative} in the main paper.
    Upper rows per inset show NeRF renderings of the original scene from novel views, with the first image also displaying the associated mask. Lower rows show the corresponding inpainted view. }
  \label{fig:additional.inpainting.qualitative}
\end{figure*}

Here, we provide additional qualitative examples to show the effectiveness of our multiview segmentation and multiview inpainting methods. Figure~\ref{fig:additional.inpainting.qualitative} is an extended version of Figure~\ref{fig:inpainting.qualitative}, and shows four additional qualitative examples of our view-consistent inpainting approach. 

\begin{figure*}[t]
  \centering
   \includegraphics[width=1.0\linewidth]{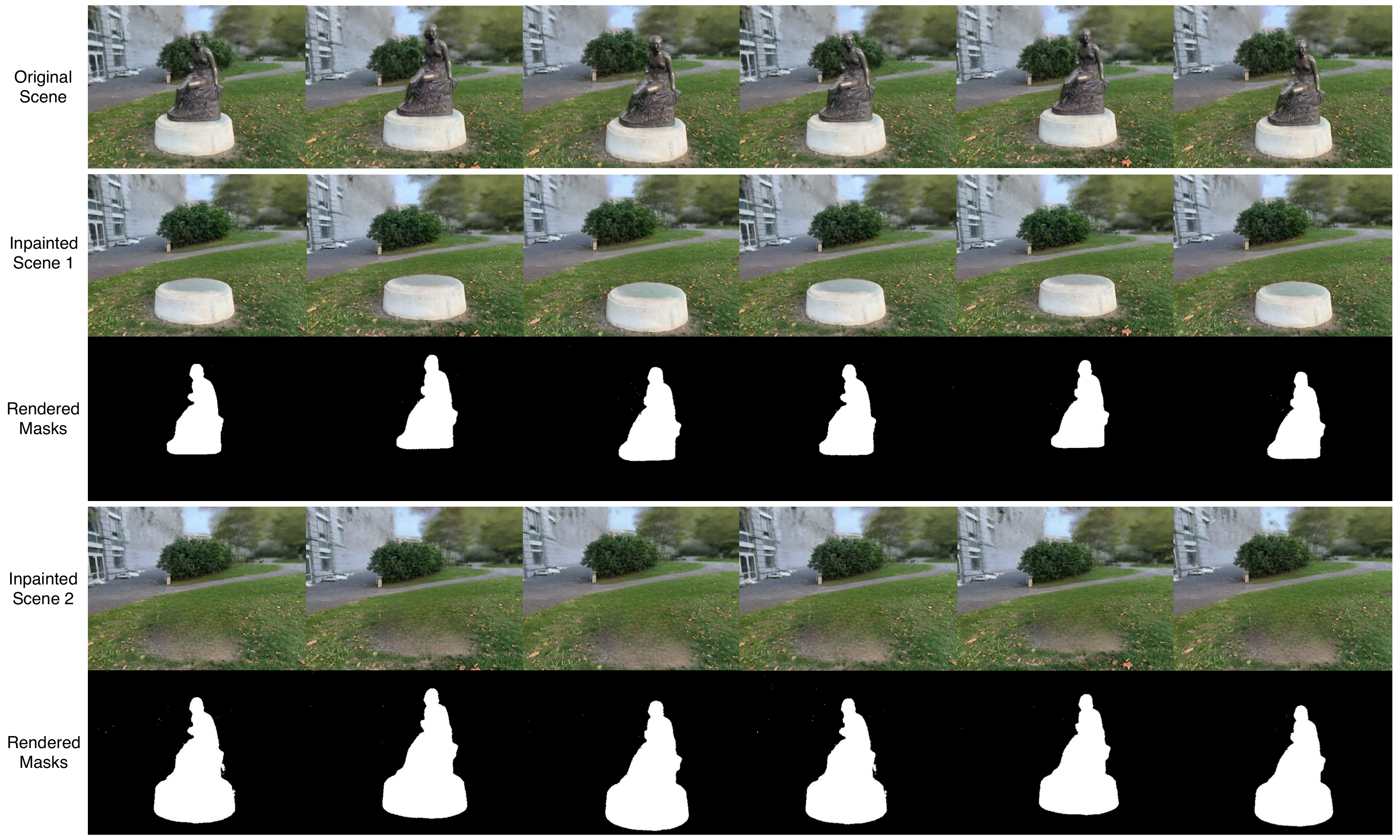}

   \caption{A single scene inpainted with two different masks using our multiview inpainting method.}
   \label{fig:diff.mask.inpainting}
\end{figure*}

Figure~\ref{fig:diff.mask.inpainting} shows an example of a single scene being inpainted twice, each time with a different part of the scene being masked. In the upper case, the statue without its concrete base is selected and the base is still in the scene after the inpainting. Notice that parts of the base as well as parts of the ground behind it were not visible in any of the training views. Our model shows consistent plausible hallucinations, which complete the cylinder shape of the base. The use of the perceptual loss leads to a sharp texture on the grass.

\begin{figure*}[t]
  \centering
   \includegraphics[width=1.0\linewidth]{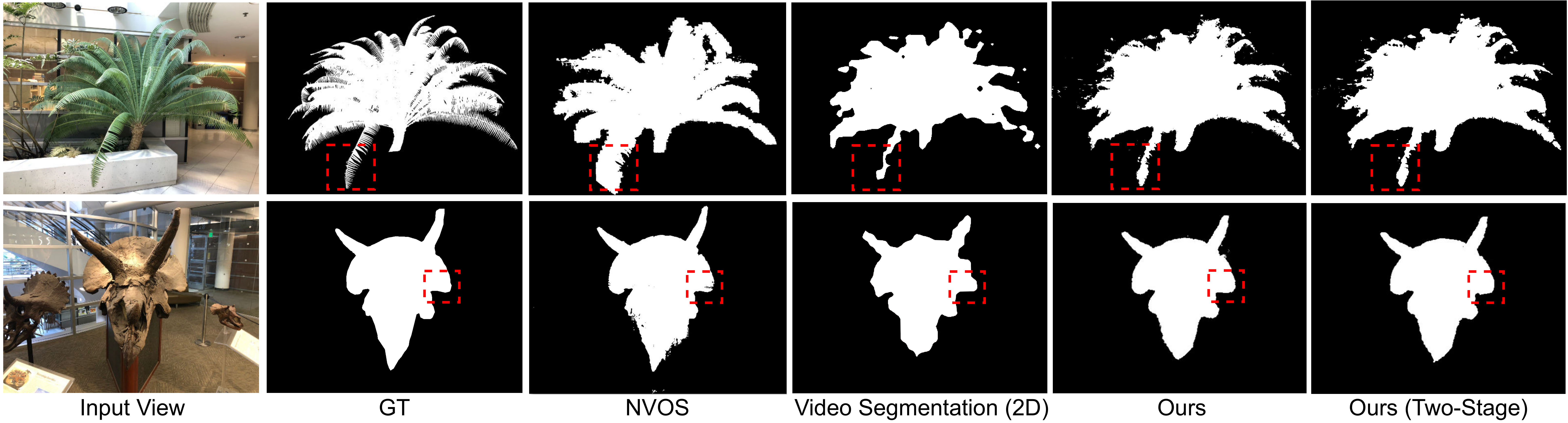}

   \caption{Qualitative comparison, as in Figure~\ref{fig:mv.seg.qualitative} in the main paper, of our multiview segmentation model against Neural Volumetric Object Selection (NVOS)~\cite{neural.object.selection}, Video segmentation~\cite{dino}, and the human-annotated masks (GT). }
   \label{fig:additional.mv.seg.qualitative}
\end{figure*}

We further provide more qualitative results of our multiview segmentation model. Figure~\ref{fig:additional.mv.seg.qualitative} is an extension of Figure~\ref{fig:mv.seg.qualitative}, and shows target views from two scenes, the ground-truth mask in the target views, and the outputs of NVOS~\cite{neural.object.selection}, video segmentation~\cite{dino}, and our model with or without the two-stage training. As evident in the results, our segmentation model consistently provides coherent masks with sharp accurate edges (zoom into boxes for examples).

\begin{figure}[t]
  \centering
   \includegraphics[width=1.0\linewidth]{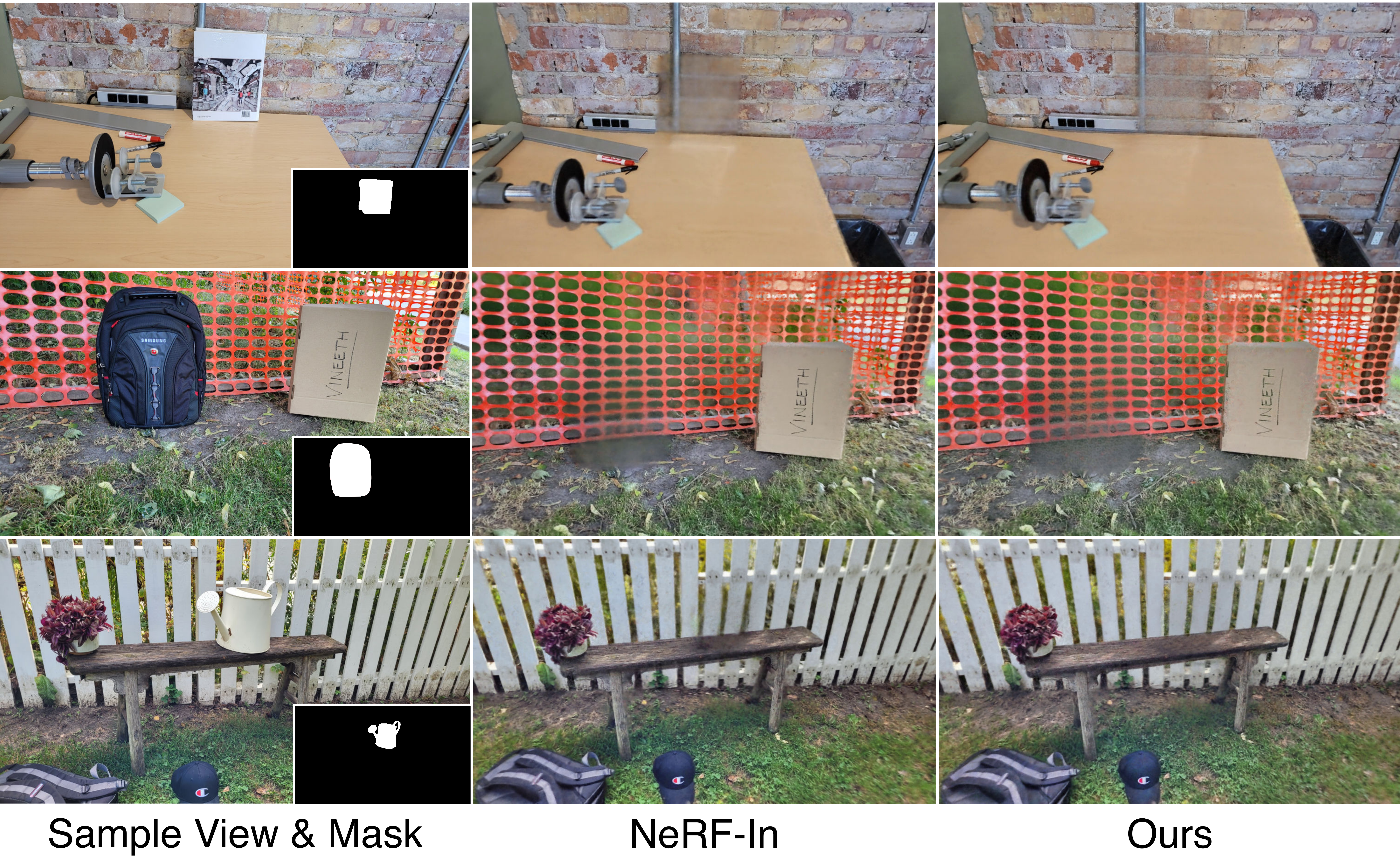}

   \caption{Additional qualitative comparisons of our model against NeRF-In~\cite{nerf.in}. }
   \label{fig:additional.comparison.inpainting.qualitative}
\end{figure}

Figure~\ref{fig:additional.comparison.inpainting.qualitative} shows additional qualitative comparisons of our model against NeRF-In~\cite{nerf.in} on three of the scenes of our dataset. As visible in the outputs, our models is able to produce sharper outputs.

\section{Multi-Stage Multiview Segmentation}\label{sec:mv.seg.multistage}

\begin{figure*}[t]
  \centering
   \includegraphics[width=1.0\linewidth]{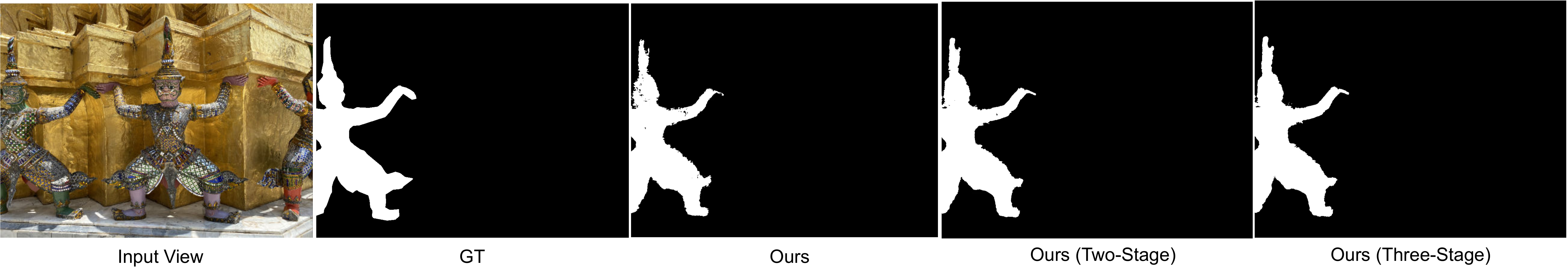}

   \caption{Qualitative comparison of our multiview segmentation model with two-stage and three-stage optimizations. As evident in the results, three-stage optimization does not lead to a significant improvement over the two-stage fitting.  }
   \label{fig:two.stage.three.stage}
\end{figure*}

While it has been shown both qualitatively (Figure~\ref{fig:mv.seg.qualitative}) and quantitatively (Table~\ref{tab:multiview.segmentation}) that our multiview segmentation benefits from our proposed two-stage training, Figure~\ref{fig:two.stage.three.stage} shows that additional training stages do not have a significant effect on the outputs, and thus, two training stages are sufficient. Quantitatively, Table~\ref{tab:multiview.segmentation.multi.stage} shows that our model with two or three stages of training has similar performance.

\section{Our Multiview Inpainting Dataset} \label{sec:mv.dataset.appendix}
\begin{figure*}[t]
  \centering
   \includegraphics[width=1.0\linewidth]{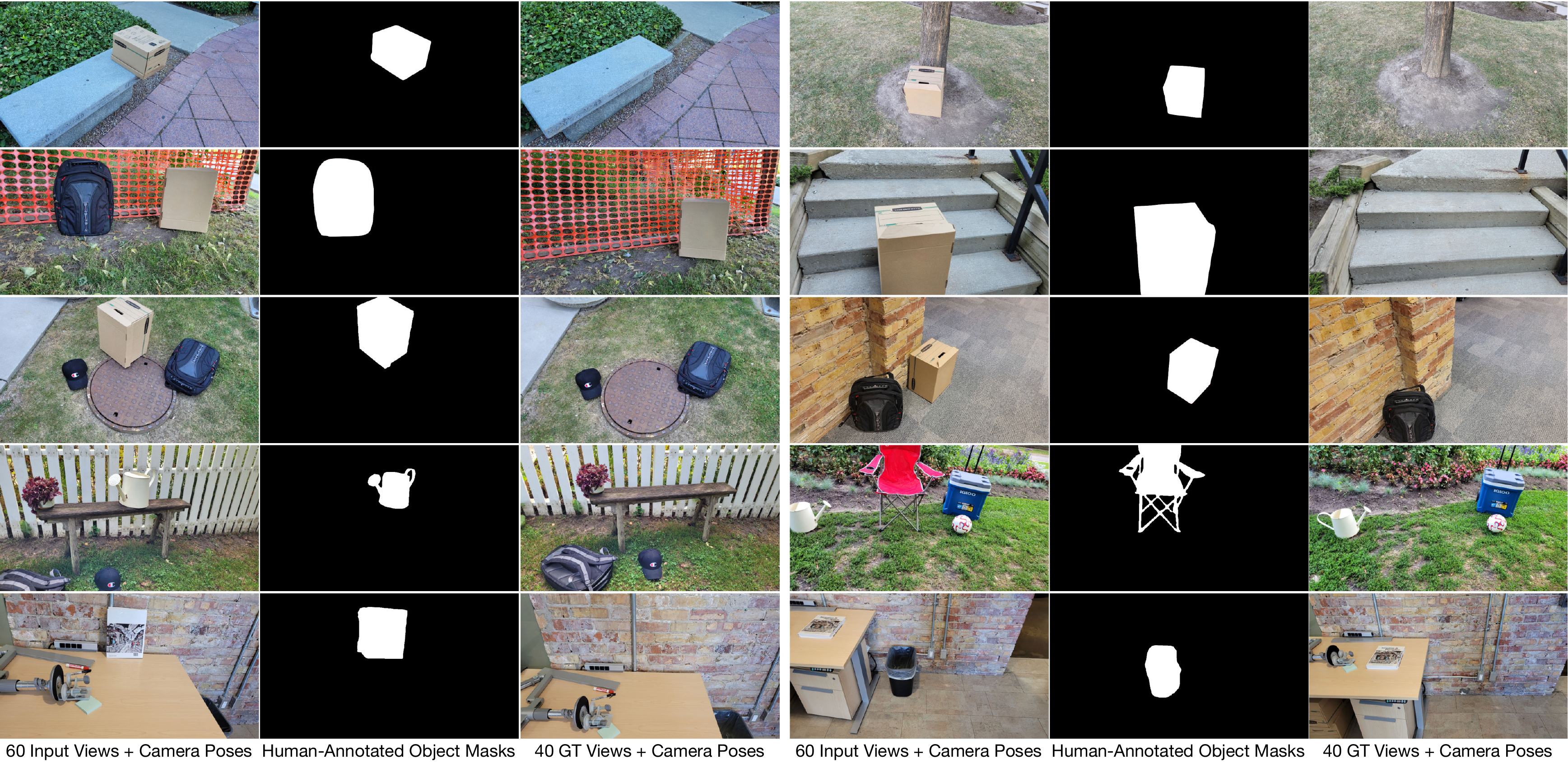}

   \caption{Overview of the 10 different scenes in our introduced dataset for multiview inpainting. }
   \label{fig:full_scenes}
\end{figure*}

Figure~\ref{fig:full_scenes} contains sample images from our introduced dataset used in our quantitative evaluations. This dataset contains 10 real-world scenes and includes different challenging 3D inpainting segmentation and inpainting scenarios. In the experiments, we use this dataset to provide a quantitative comparison of our inpainting method against the baselines, where our approach outperforms other methods.

\begin{table}[tb]
\centering
\caption{Quantitative evaluation of our proposed multiview segmentation with one, two, and three training stages. }
\begin{tabular}{lcc}
\hline
\multicolumn{1}{l}{\textbf{$\#$ of Stages}} & \textbf{Acc.}$\uparrow$ & \textbf{IoU}$\uparrow$ \\ \hline
1                                               & 98.85                         & 90.96              \\
2                                               & \textbf{98.91}                & \textbf{91.66}  \\
3                                               & 98.89                         & 91.53              \\
\hline      
\end{tabular}
\label{tab:multiview.segmentation.multi.stage}
\end{table}

\section{NeRF: Extended Background} \label{sec:extended.bg}

Here, we provide an extended version of the background on Neural Radiance Fields (NeRFs) for completeness. NeRFs~\cite{original.nerf} encode a 3D scene as a function, $f:(x, d) \rightarrow (c, \sigma)$, that maps a 3D coordinate, $x$, and a view direction, $d$, to a color, $c$, and density, $\sigma$. The function $f$ can be modelled in various ways, such as a multilayer perceptron (MLP) with positional encoding~\cite{original.nerf} or a discrete voxel grid with trilinear interpolation~\cite{plenoxels}, depending on the application and desired properties. For a 3D ray, $r$, characterized as $r(t) = o + td$, where $o$ denotes the ray's origin, $d$ its direction, and $t_n$ and $t_f$ the near and far bounds, respectively, the expected color is:
\begin{equation}
    \label{eq:volumetric.rendering2}
    C(r) = \int_{t_n}^{t_f} T(t) \sigma(r(t))c(r(t), d) \diff t, 
\end{equation}
where $T(t) = \exp(- \int_{t_n}^{t} \sigma(r(s)) \diff s)$ is the transmittance. The integral in Eq.~\ref{eq:volumetric.rendering2} is estimated via quadrature by dividing the ray into $N$ sections and sampling $t_i$ from the $i$-th section:
\begin{equation}
    \label{eq:volumetric.rendering.discrete2}
    \hat{C}(r) = \sum_{i = 1}^{N} T_i(1 - \exp(-\sigma_i \delta_i))c_i,
\end{equation}
where $T_i = \exp(-\sum_{j = 1}^{i - 1} \sigma_j \delta_j)$ and $\delta_i = t_{i + 1} - t_i$ is the distance between two adjacent sampled points. For simplicity, $c(r(t_i), d)$ and $\sigma(r(t_i))$ are abbreviated as $c_i$ and $\sigma_i$, respectively. For the rays passing through pixels of the training views, the ground-truth color, $C_{\text{GT}}(r)$, is available, and the representation is optimized using the reconstruction loss:
\begin{equation}
    \label{eq:reconstruction.loss2}
    \mathcal{L}_\text{rec} = \sum_{r \in \mathcal{R}} \Vert \hat{C}(r) - C_{\text{GT}}(r) \Vert ^2,
\end{equation}
where $\mathcal{R}$ is a ray batch sampled from the training views.

\section{Detailed Segmentation Results}\label{detailed.segmentation.results}

Table~\ref{tab:multiview.segmentation.expanded} shows a breakdown of Table~\ref{tab:multiview.segmentation} based on forward-facing and $360^\circ$ scenes. The inputs to all of the models in this experiment is a single-view mask, which is to be transferred to other views. As a result, the task is more challenging for $360^\circ$ scenes, due to the need to extrapolate the single-view mask to further views. 
Regardless of the differences in difficulty, our model consistently outperforms the baselines in both forward-facing and $360^\circ$ scenarios (Table~\ref{tab:multiview.segmentation.expanded}). 

\begin{table}[tb]
\centering
\vspace{-0.1 cm}   
\caption{Quantitative multi-view segmentation evaluation for forward-facing and $360^\circ$ scenes. See also Table~\ref{tab:multiview.segmentation}. }
\vspace{-0.3 cm}   
\resizebox{0.48\textwidth}{!}{
\begin{tabular}{lcccl}
\hline
       & \multicolumn{2}{c}{\textbf{Forward-Facing}} & \multicolumn{2}{c}{\textbf{360$^{\circ}$}} \\
       & \textbf{Acc.}$\uparrow$     & \textbf{IoU}$\uparrow$     & \textbf{Acc.}$\uparrow$             & \textbf{IoU}$\uparrow$            \\ \hline
Proj. + Grab Cut (2D)                  & 92.19                         & 59.84            &89.54     &28.09    \\
Proj. + EdgeFlow (2D)                  & 97.63                         & 87.00            &95.73     &74.10    \\ \hdashline
Semantic NeRF (only source mask)       & 98.72                         & 90.96            &88.90     &52.98    \\
Proj. + EdgeFlow + Semantic NeRF       & 98.74                         & 91.53            &95.20     &73.35    \\
Feature Field Distillation             & 98.20                         & 85.61            &96.19     &79.51    \\
Video Segmentation                     & 98.87                         & 91.38            &97.81     &84.08    \\
Ours (two-stage)                       & \textbf{99.29}                & \textbf{94.64}   &\textbf{98.37}     &\textbf{87.48}    \\ \hline  
\vspace{-1.2 cm}         
\end{tabular}
\label{tab:multiview.segmentation.expanded}
}
\end{table}

\section{Failure Cases}\label{failure.cases}

Since SPIn-NeRF is based on an underlying NeRF and a 2D inpainter, it is prone to the failure cases of these models; e.g., the image inpainter failing results in the failure of SPIn-NeRF as well. Moreover, despite the effectiveness of the perceptual loss in handling texture-level inconsistencies between the image priors, potential semantic-level inconsistencies can result in failure. For instance, if some inpainted views contain novel inserted objects in the masked region (in contrast to simply extending the background to remove the unwanted object, as our method expects), the perceptual loss might fail to converge to a meaningful solution. 
In particular, as the resulting independently inpainted patches would not reside nearby in the perceptual metric space, the NeRF output (attempting to balance between them) in the masked area would likely be blurry or contain other artifacts. 
Due in part to this consideration, we utilize LaMa~\cite{lama} as our underlying inpainter, as it reduces the likelihood of this scenario, since LaMa is not a ``creative'' inpainter and typically only removes objects. However, such problematic cases are likely with more creative inpainters, such as non-deterministic denoising diffusion-based inpainters.

\section{Ethics Statement} \label{sec:ethical.statement}
There has been a constant debate about 2D generative models and image manipulation techniques, and the concerns regarding potential misuses. The majority of these concerns also apply to the new line of 3D generation and manipulation~\cite{poole2022dreamfusion}. In the hands of an adversary, these models can be utilized to manipulate people's perception of reality and generate disinformation. Moreover, the fact that LaMa~\cite{lama} is used in our implementation results in the inheritance of LaMa's potential undesirable biases in the outputs of our 3D inpainter.

\end{document}